\documentclass[times,twocolumn,final]{elsarticle}

\usepackage{geometry}
\usepackage{framed,multirow}
\usepackage{afterpage}

\usepackage{amssymb}
\usepackage{latexsym}

\usepackage{url}
\usepackage[dvipsnames]{xcolor}

\usepackage{hyperref}
\usepackage{caption} 

\definecolor{newcolor}{rgb}{.8,.349,.1}

\journal{ }

\begin{document}

\newgeometry{left=1.3in,right=1.3in,top=2in,bottom=1in}

\begin{frontmatter}

\title{HyperFusion: A Hypernetwork Approach to Multimodal Integration of Tabular and Medical Imaging Data for Predictive Modeling}

\author[1]{Daniel Duenias}
\author[2,3]{Brennan Nichyporuk}
\author[2,3]{Tal Arbel}
\author[1]{Tammy Riklin Raviv}

\author[]{ADNI \corref{cor1}}
\cortext[cor1]{Data used in preparation of this article were obtained from the Alzheimer’s Disease Neuroimaging Initiative (ADNI) database (adni.loni.usc.edu). As such, the investigators within the ADNI contributed to the design and implementation of ADNI and/or provided data but did not participate in analysis or writing of this report. A complete listing of ADNI investigators can be found at: \url{http://adni.loni.usc.edu/wp-content/uploads/how_to_apply/ADNI_Acknowledgement_List.pdf}}

\address[1]{Ben Gurion University of the Negev, blvd 1, Beer Sheva 84105, Israel}
\address[2]{Centre for Intelligent Machines, McGill University, 3480 University St, Montreal, QC, H3A 0E9, Canada}
\address[3]{Mila - Quebec AI Institute, 6666 Rue Saint-Urbain, Montréal, QC H2S 3H1, Canada}

\begin{abstract}
The integration of diverse clinical modalities such as medical imaging and the tabular data extracted from patients' Electronic Health Records (EHRs) is a crucial aspect of modern healthcare.
Integrative analysis of multiple sources can provide a comprehensive understanding of the clinical condition of a patient, improving diagnosis and treatment decision. Deep Neural Networks (DNNs) consistently demonstrate outstanding performance in a wide range of multimodal tasks in the medical domain. However, the complex endeavor of effectively merging medical imaging with clinical, demographic and genetic information represented as numerical tabular data remains a highly active and ongoing research pursuit.

We present a novel framework based on hypernetworks to fuse clinical imaging and tabular data by conditioning the image processing on the EHR's values and measurements. This approach aims to leverage the complementary information present in these modalities to enhance the accuracy of various medical applications. We demonstrate the strength and generality of our method on two different brain Magnetic Resonance Imaging (MRI) analysis tasks, namely, brain age prediction conditioned by subject's sex and multi-class Alzheimer's Disease (AD) classification conditioned by tabular data. We show that our framework outperforms both single-modality models and state-of-the-art MRI tabular data fusion methods. A link to our code can be found at \href{https://github.com/daniel4725/HyperFusion}{https://github.com/daniel4725/HyperFusion}. 
\end{abstract}

\begin{keyword}
 Deep Learning \sep Hypernetworks\sep Tabular-Imaging Data Fusion\sep Multimodal \sep  Alzheimer Diagnosis \sep Brain Age Prediction
\end{keyword}

\end{frontmatter}


\section{Introduction} \label{sec:intro}
In medical decision-making, clinicians rely on comprehensive patient data, including clinical, genetic, demographic, and imaging information, to gain an in-depth understanding of an individual's condition, thereby improving diagnostic accuracy and treatment outcomes.
Deep Neural Networks (DNNs) consistently achieve state-of-the-art results in a broad spectrum of medical tasks. However, they still fall short of matching human experts' abilities to integrate information from medical imaging and non-imaging data.
Images are high-dimensional, continuous, and spatial, while patients' Electronic Health Records (EHRs), formatted as numerical, tabular data, typically comprise low-dimensional quantitative attributes with diverse types, scales, and ranges. \restoregeometry \noindent Vision-Language Models (VLMs), such as CLIP (Contrastive Language–Image Pretraining) \citep{radford2021CLIP}, have demonstrated remarkable success in capturing cross-modal relationships between textual concepts and images. These methods rely on contrastive learning, where the model is trained to align representations of images with their closely related textual descriptions while distinguishing unrelated pairs. However, in the medical domain, challenges arise due to the non-uniform distribution of data \citep{wang2023sample}. Specifically, there are often no explicit, one-to-one semantic mappings; the same image can be associated with multiple concepts, and a single concept may correspond to different images.

The fusion of imaging and tabular (rather than textual) data presents an additional complication. Although certain measures, such as size or morphology, can be directly correlated with scanned anatomy, many other clinical and demographic attributes, such as medical laboratory results or patient demographics, offer complementary information independent of imaging data. 
There is no "correct" or semantic match between a patient's chest X-ray and their body temperature; the latter, being a continuous variable, can take any value within a given range. However, when combined, both sources of data can be used to improve the assessment of conditions such as pneumonia~\citep{Huang2020survey}.

To fully exploit both visual and tabular information, we present a conceptually innovative approach that treats clinical measurements and demographic data as priors, influencing the outcomes of an image analysis network. The entire fusion process is analogous to tuning the auditory output of a radio transmitter by manipulating the frequency (selecting stations) and volume. This central concept is implemented through hypernetworks. Originally introduced by \citep{ha2016hypernetworks}, the hypernetwork framework consists of a primary network assigned to a specific task and a hypernetwork that generates the weights and biases for specific layers of the primary network. The main strength of this meta-learning framework is that, unlike standard deep models, which remain fixed after training, the primary image-processing network is dynamically adjusted based on the input tabular attributes, even at test time.

The key contributions of this work include: (1) the introduction of HyperFusion, a novel hypernetwork-based framework that effectively fuses imaging and tabular data, where tabular information serves as priors influencing the outcomes of an image analysis network. This dynamic fusion process is shown to be particularly powerful, allowing the image-processing network to be adjusted at test time based on varying input attributes. (2) The application of this framework to two challenging brain MRI analysis tasks: brain age prediction conditioned by sex and classification of Alzheimer's disease (AD), Mild Cognitive Impairment (MCI), and Cognitively Normal (CN) subjects using their clinical, demographic, and genetic data. (3) Empirical results demonstrating that HyperFusion outperforms state-of-the-art methods in both tasks, providing new insights into the integration of imaging and tabular data in medical diagnostics.

The rest of this paper is organized as follows. Section \ref{sec:related_work} reviews the fusion of imaging and tabular data, including the concept of hypernetworks and their applications, along with relevant studies on brain age prediction and AD classification. Section \ref{sec:method} introduces our hypernetwork framework, detailing the architectures for each task and key implementation aspects. In Section \ref{sec:experiments}, we describe the experimental setup, including datasets, preprocessing, and training/evaluation procedures, followed by the presentation and discussion of the results. Finally, we conclude in Section~\ref{sec:discussion_conclusions}.


\section{Related Work} \label{sec:related_work}

\subsection{Fusion methods} 
\textcolor{black}{The integration of multiple, diverse medical data modalities is an active area of study~\citep{heiliger2023beyond}. While the fusion of multimodal imaging datasets has been explored for over a decade~\citep{menze2014multimodal, carass2017longitudinal, sui2023data}, the merging of imaging and non-imaging data presents additional challenges, and relevant algorithms are relatively recent.}

Fusion strategies for combining medical imaging and tabular data can be categorized into three distinct types, as outlined in~\citep{Huang2020survey}: early fusion, where original or extracted features are concatenated at the input level, e.g.,~\citep{chieregato2022COVID_CTandEHR}; joint fusion (or intermediate fusion), where the feature extraction phase is learned as part of the fusion model; and late fusion, where predictions or pre-trained high level features are combined at the decision level, as in~\citep{pandeya2021late_fusion_softmax, prabhu2022multimodal_AD_entropydecision}.

In this study, we employ the joint fusion approach to facilitate meaningful interaction between modalities. Due to the inherent disparities between tabular and image data, direct integration is not feasible, necessitating some preprocessing. Early fusion proves inadequate for end-to-end processing, leading to the independent processing of image features and hindering the potential for mutual learning between modalities at the intermediate level. Late fusion, on the other hand, occurs solely at the decision level of trained models, failing to foster mutual learning between modalities. In contrast, joint fusion offers the advantages of end-to-end training as well as the potential to condition modality processing based on each other.

A straightforward and intuitive approach to fuse image and tabular data involves using separate networks for each data type - typically a Multi-Layer Perceptron (MLP) for tabular data and a Convolutional Neural Network (CNN) for imaging.
While late fusion methods provide predictions based on collective decisions~\citep{prabhu2022multimodal_AD_entropydecision} - joint fusion frameworks embed each modality into a feature vector followed by vector concatenation \citep{esmaeilzadeh2018concat_fusing, el_appagh2020Multimodalmultitask, venugopalan2021ADfusion_img_tab_snp}. However, concatenation methods limit the interaction between tabular data and imaging to high-level descriptors, neglecting fusion in CNN layers where crucial spatial image context is preserved.

More advanced joint fusion techniques apply affine transformation to the image features
within the intermediate CNN layers, where the shift and scale parameters are generated by a network that processes another data modality.   
In~\citep{perez2018film}, a Feature-wise Linear Modulation (FiLM) was introduced to condition image processing on text, applied to a visual question-answering task. Building upon FiLM, \citep{wolf2022DAFT} introduced the Dynamic Affine Feature Map Transform (DAFT) to fuse brain MRIs and tabular data. However, in these techniques, the interaction between different data types is confined to specific stages of the processing pipeline and linear transformations, potentially limiting the network's ability to fully exploit the advantages of data fusion.

Addressing a different, yet related problem, \citep{Hager2023contrastive_cvpr} tackles the scenario where tabular data is assumed to be unavailable during the inference phase but can be utilized during training to leverage semantic dependencies within imaging data. The methodology employs contrastive learning to understand imaging-tabular correspondences. However, as highlighted in Section \ref{sec:intro}, our work observes that the most relevant tabular features are distinctly present in the imaging data. In certain cases, the tabular data encompasses vital attributes that are challenging or nearly impossible to derive from imaging alone, such as specific protein measurements and metabolic indicators. Moreover, it is worth noting that \citep{Hager2023contrastive_cvpr} relies on extensive datasets for effective contrastive learning utilization, while our work also addresses situations with smaller datasets lacking comprehensive representation. In such instances, explicitly providing information from the tabular data can prove more beneficial than relying on implicit extraction from the image.

Our work incorporates the concept of conditioning image processing on tabular data by employing a hypernetwork framework. 
In contrast to other methods, our hypernetwork approach is versatile and comprehensive. Specifically, the proposed hypernetwork learns a general transformation (not necessarily linear) to adapt the parameters transferred to the primary network's layers, regardless of their architecture or position. This flexibility allows for the fusion of imaging with any other modality regardless of the context.  

\subsection{Hypernetworks}
The notion of hypernetworks, where one network (the hypernetwork) generates the weights and biases for another network (the primary network), was introduced in~\citep{ha2016hypernetworks}. We took inspiration from two distinct hypernetwork applications, one of which is the work of \citep{littwin2019hyper3Dreconst}. This work utilized 2D images input into a hypernetwork for the reconstruction of depicted 3D objects via the primary network. Another inspiring application is found in the denoising framework presented by~\citep{aharon2023}, where the hypernetwork was trained to generate primary network parameters conditioned by the latent signal-to-noise ratio (SNR) of the input image.

Hypernetworks have also found applications in the medical domain. In \citep{wydmanski2023hypertab}, a hypernetwork framework processes tabular data using random feature subsets. In this context, a random binary mask selects a subset of tabular features, simultaneously serving as the input to the hypernetwork responsible for generating parameters for the feature processing network. However, it is important to note that this approach is unimodal, focusing exclusively on tabular data, and lacks integration with imaging data, a key aspect addressed in our work.

To the best of our knowledge, the use of hypernetworks for the fusion of different data modalities in the medical domain is proposed here for the first time.
To showcase our approach, we selected two brain MRI analysis challenges that can potentially benefit from the fusion of imaging and tabular data: brain age prediction, a regression problem, and multi-class classification, involving categorization into AD, MCI and CN. More details about these tasks are provided below.

\subsection{Brain age prediction}
The task of predicting human age based on brain anatomy has garnered significant attention in medical research, driven by the development of AI regression tools and the increasing availability of brain imaging data~\citep{franke2019BrainAGE10years, Peng2021lightDNN4BrainAge, lee2022BrainAgePrediction, cole2017brainage_biomarker, feng2020BrainAgeDNN, levakov2020brainage}.
Current methods, exclusively based on imaging, demonstrate good prediction accuracy, achieving an average gap of less than three years between the actual chronological age and the predicted brain age of healthy subjects. However, since the primary goal of these studies is the explainability of the neural network, with the aim of providing information on the dynamics of brain aging, relying solely on imaging does not capture other attributes that may influence these processes.

Recognizing the differences in the aging trajectories between men and women~\citep{coffey1998AgeAndSex,piccarra2023}, we explored whether the incorporation of subject sex information could improve the prediction of brain age. To the best of our knowledge, we are the first to address the task of brain age inference conditioned by the subject's sex. However, it is essential to note that this exploration is a secondary objective of our study. The primary objective is to showcase and demonstrate our proposed hypernetwork framework, specifically focusing on the fusion of imaging and nonimaging data.

\subsection{multi-class AD classification}
Early detection of AD from brain MRI is a topic of active research due to its significant implications. The most challenging aspect lies in the classification of individuals with MCI, a transition stage that serves as a potential predictor of the development of AD. Unlike methods designed for binary classification (CN vs AD), such as those presented in \citep{backstrom2018CN_ADclass, wang2018CN_ADclass}, our focus is on multiclass classification into CN, MCI and AD.

Although many studies addressing this problem rely only on imaging data, as seen in \citep{wen2020CNN_AD_overview} and related works, the literature related to neurology suggests the inclusion of clinical and demographic data for enhanced prediction. For example, works such as \citep{Letenneur1999SexAgeEducation_AD, Fratiglioni1886AD_and_education} highlight differences in the hazard ratio of AD between women and men, as well as among individuals with varying levels of educational attainment. The Alzheimer's Disease Neuroimaging Initiative (ADNI) dataset offers additional tabular data containing essential attributes and biomarkers for AD detection, such as specific protein measurements in the Cerebrospinal Fluid (CSF) and metabolic indicators derived from Positron Emission Tomography (PET) scans.

These additional attributes and their interactions provide relevant information that may not be present in brain MRI scans alone. In recent years, several multimodal fusion methods have been proposed for AD detection~\citep{Dolci2022, venugopalan2021ADfusion_img_tab_snp, el_appagh2020Multimodalmultitask, zhou2019ADfusion_MRI_PET_genetic, liu2018ADfusion_concat, Simeon2018_2classADfusion}.
In this manuscript, we specifically refer to \citep{wolf2022DAFT, esmaeilzadeh2018concat_fusing, prabhu2022multimodal_AD_entropydecision}, which addressed the fusion of MRIs and tabular data from the ADNI dataset for three-class AD classification.

\section{Method}  \label{sec:method}

\subsection{HyperFusion}  \label{sec:method_our_hypernetwork_framework}
\begin{figure}[!t]
    \centering
    \includegraphics[scale=0.65]{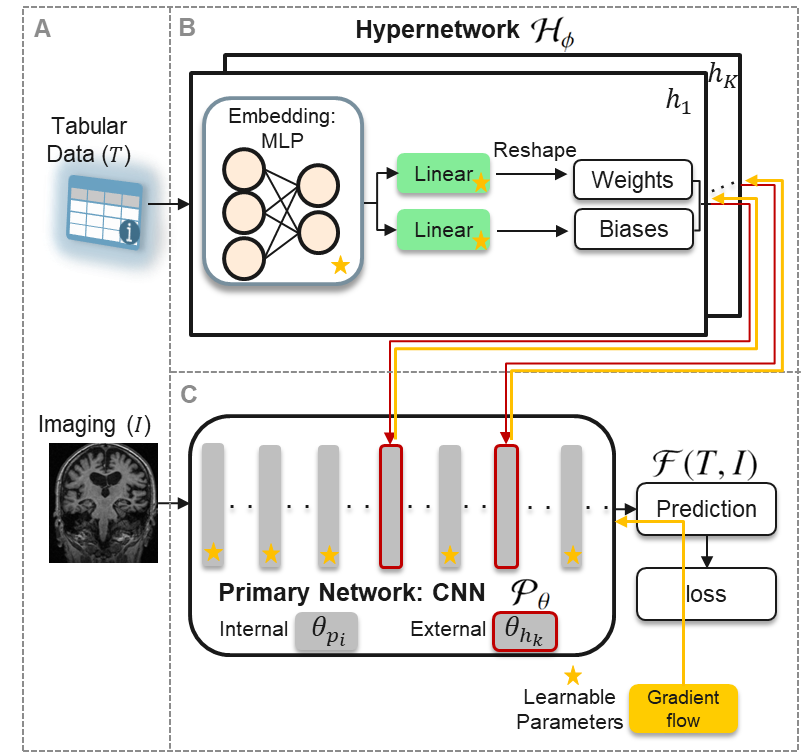}
    \caption{
    \textcolor{black}{An illustration of the proposed HyperFusion's Framework.
     The two main components - hypernetwork and primary network are shown in the upper and the lower part of the figure, respectively. A: The inputs $T$ and $I$ denote tabular and imaging data, respectively. B: The hypernetwork $\mathcal{H}_\phi$ is composed of $K$ individual networks, $\{h_k\}_{k=1, \ldots, K}$, which generate parameters $h_k(T)=\theta_{h_k}$ for specific (external) layers of the primary network $\mathcal{P}_\theta$ (red arrows). 
    C: The primary network is composed of internal layers which are updated throughout the backpropagation process (yellow arrows) and external layers (marked in red).}
    \label{fig:our_basic_hyper}} 

\end{figure}

The proposed deep learning framework aims to integrate imaging and tabular data to enhance clinical decision making. Figure~\ref{fig:our_basic_hyper} describes its two main building blocks: a hypernetwork denoted by $\mathcal{H}_\phi$ and a primary network $\mathcal{P}_\theta$, where $\phi$ and $\theta$ represent their respective parameters. For simplicity, we denote the entire network compound by $\mathcal{F} = \{\mathcal{H}_\phi, \mathcal{P}_\theta\}$.

The input to $\mathcal{F}$ consists of a paired tabular vector of $d$ measurements/values $T \in \mathbb{R}^d$ and a multidimensional image $I$, specifically a 3D MRI (Figure~\ref{fig:our_basic_hyper}A). Given a tabular vector $T$, $\mathcal{H}$ produces a data-specific set of network parameters $\theta_\mathcal{H} = \mathcal{H}_\phi(T)$, which, combined with internally learned parameters $\theta_\mathcal{P}$, forms the entire set of primary network parameters $\theta = \{\theta_\mathcal{H}, \theta_\mathcal{P}\}$. The output of the primary network $\mathcal{P}_\theta(I)$ represents the desired clinical prediction. 
In the feedforward pass the input image is fed into the primary network $\mathcal{P}_\theta$ (Figure~\ref{fig:our_basic_hyper}C). Simultaneously, the tabular data input is processed by the hypernetwork $\mathcal{H}_\phi$ (Figure~\ref{fig:our_basic_hyper}B). The hypernetwork block is 
composed of $K$ sub-networks $\{h_k\}_{k=1, \ldots, K}$, each embeds the tabular attributes (see Section~\ref{sec:method_embedding}), and generates weights and biases $h_k(T) = \theta_{h_k}$ for a corresponding layer in the primary network $\mathcal{P}_\theta$ (framed in red). These external parameters of the primary network's layers depend on $T$ and influence the extracted imaging features accordingly.

During the training phase, the loss backpropagates (yellow arrows) through both the primary and hypernetwork, influencing the learnable parameters $\theta_\mathcal{P}$ and $\phi$ (marked with yellow stars). Specifically, the gradients update the internal primary network layers and only pass through the external ones to update the parameters in the respective hypernetworks. 
Notably, the external primary network parameters $\theta_\mathcal{H}$ are updated indirectly by the hypernetwork (red arrows). 
This structure allows us to condition the image analysis on the tabular data.

The dependency on tabular data is determined by the ratio between external and internal parameters. To gain insights, let us examine two edge cases. When relying solely on internal network parameters (without a hypernetwork), predictions become indifferent of the tabular information. Conversely, the case where all primary network parameters are external and generated by the hypernetwork, approximates a scenario where separate networks are employed for each combination of tabular attributes. For instance, in brain age prediction, the hypernetwork might generate different parameter sets for men and women. However, while there will be some dependencies thanks to the embedding (Section~\ref{sec:method_embedding}), having all parameters external would probably necessitate an extended training process with a higher amount of training data. In our configurations, we opted for parameters of low-level primary network layers to be internal, assuming that tabular attributes (e.g., sex) are less likely to be relevant for the extraction of the low level imaging features that eventually contribute to the prediction. 
This chosen combination of parameter sources allows us to maximize the utilization of available training scans while incorporating the corresponding tabular data for additional gains.

\subsubsection{Embedding the tabular data}  \label{sec:method_embedding}
As depicted in Figure \ref{fig:our_basic_hyper}B, the proposed hypernetwork generates weights and biases in a two-step process. Initially, the tabular data is input into an embedding network, resulting in a latent vector. Subsequently, this latent vector passes through two distinct linear layers: one for generating weights and another for generating biases.
Let $\zeta\colon\mathbb{R}^d \to \mathbb{R}^l$ define the embedding function, mapping the tabular data $T$ 
into a lower dimension space ($l < d$). The desired embedded feature vectors 
should have a condensed, meaningful representation of the original data, aiming to capture hidden patterns and intricate interactions, while preserving the relevant information. The embedding is a critical step in the fusion process, offering complex associations to the tabular data.

For our tabular embedding network, we opted for an MLP model given that tabular data lacks spatial context. This choice aligns with its favorable performance and suitability for small datasets, as noted in \citep{borisov2022surveyDNN_tabular}. However, the hypernetwork configuration remains versatile, allowing the integration of any desired embedding architecture. 

The proposed framework is end-to-end and the embedding parameters are learned concurrently with the entire network, as illustrated in Figure~\ref{fig:our_basic_hyper}.
Moreover, the continuity of the embedding space, where similar samples are mapped closer together, enables interpolation. This capability allows the model to generate meaningful representations for previously unseen samples, enhancing its ability for generalization

\subsubsection{Hyperlayer's position selection}  \label{sec:hyperlayers-selection}
While the proposed hypernetwork concept is general, its adaptation to different applications requires the accommodation to specific architectures of the primary network. 
Inspired by the layer selection method proposed in ~\citep{lutati2021hyperhyper}, we identify `good candidates' for hyperlayers by evaluating the impact of each primary network layer given the untrained, backbone network.  The evaluation is carried out by randomly initializing the parameters of the selected layer multiple times ($N=1000$) while keeping the random parameters of the other network layers fixed. 
The entropy of the layer-based loss is calculated based on the normalized histogram of the loss values obtained for each random initialization of the layer of interest. Final layer selection from the pool of candidates is conducted empirically as demonstrated in the ablation study in~\ref{sec:hyperlayers-selection}. 

\subsubsection{Weights initialization} \label{sec:weights-initialization}
A network's convergence is contingent upon the initial distribution of its parameters. Typically, neural network layer parameters are initialized taking into account their fan-in or fan-out, a practice that stabilizes their convergence, as outlined in~\citep{he2015Kaiming_init, glorot2010xavier_init}. In our context, parameter initialization assumes a crucial role, particularly concerning $\theta_\mathcal{H}$ generated by the hypernetwork (its output) for the primary network. These parameters are computed as part of the hypernetwork's feedforward process and cannot be directly initialized to follow a specific distribution. Furthermore, these parameters vary with each different input sample $T$.

To address this challenge, we employ the technique detailed in~\citep{chang2019hyper_weights_init}, which performed variance analysis on the hypernetwork output, $\theta_\mathcal{H}$. The method aims to initialize the parameters $\phi_k$ of each hypernetwork $k$ in a way that enables the hypernetwork feedforward process to generate external primary network parameters ($\theta_{h_k}$) fitting a specific distribution. More precisely, the distribution of these parameters should ensure that the output variance in their designated layers within the primary network is closely aligned with the input variance. This initialization contributes to the convergence of the proposed hypernetwork framework.

\textcolor{black}{Formally, let $\phi_k =\{H^W_k, H^B_k\}$ denote the parameters of the fully connected layers of $h_k$ that respectively generate the weights $W^j_k \in \theta_{h_k}$ and the biases $B^j_k \in \theta_{h_k}$ to a hyperlayer $j$ in the primary network. 
Let $d_k$ and $d_j$ denote the fan-in size of the hypernetwork and the primary network layers, respectively. We also denote by $e(T)$ the embedding (the output of the MLP) of the pre-processed tabular data and by $\mbox{Var}(\cdot)$ the variance.
To achieve the desired variance, the parameters $H^M_k$ of each hypernetwork $k$ are initialized by sampling from a uniform distribution between $-\sqrt{3\mathcal{V}(H^M_k)}$ and $\sqrt{3\mathcal{V}(H^M_k)},$ where $M$ either stands for the weights $W$ or biases $B$ and $\mathcal{V}(H^M_k)$ is calculated for each set of parameters, as in~\citep{chang2019hyper_weights_init} as follows:}
\textcolor{black}{\begin{equation}
    \mathcal{V}(H^W_k) = \frac{1}{d_j \cdot d_k \cdot \mbox{Var}(e(T))}
\end{equation}
\begin{equation}
    \mathcal{V}(H^B_k) = \frac{1}{d_k \cdot \mbox{Var}(e(T))}
\end{equation}}
\subsection{Loss functions} \label{sec:methods_loss}
As depicted by the yellow arrows in Figure~\ref{fig:our_basic_hyper} the loss backpropagates throughout the entire network compound affecting both the internal primary network's parameters and the hypernetwork.
Given the flexibility of our framework, the loss function can be adapted to a wide range of applications.
It consists of two components the task-specific loss, $\mathcal{L}_{\mbox{\footnotesize{task}}}$, and a weight decay regularization term $\mathcal{L}_{\mbox{\footnotesize regularization}}$ as follows:
\begin{equation} \label{eq:loss_func}
    \mathcal{L}(y, \mathcal{F}(T, I)) = \mathcal{L}_{\mbox{\footnotesize{task}}}(y, \mathcal{F}(T, I)) + \mathcal{L}_{\mbox{\footnotesize {regularization}}}(\{\phi, \theta_{\mathcal{P}}\}),
\end{equation}
where, $y$ are the true labels/values.
Regularization is applied to $\phi$ and $\theta_{\mathcal{P}}$ to reduce overfitting. Recall that the external primary network parameters produced by the hypernetwork, i.e., $\theta_{\mathcal{H}}$ do not go through the gradient descent process and therefore are not regularized.

The loss term $\mathcal{L}_{\mbox{\footnotesize{task}}}$ in regression tasks is the Mean Square Error (MSE) between the predictions $\hat{y}$ and the ground truth values as follows:
\begin{equation} \label{eq:loss_regression}
    \mathcal{L}_{\mbox{\footnotesize{task = regression}}}(y, \hat{y}) = \frac{1}{B} \sum_{i=1}^B (y_i - \hat{y}_i)^2
\end{equation}
where $B$ is the batch size and $i$ is the sample index.

For classification tasks with $C$ classes, in order to account for possible class imbalance, we use the Weighted Cross-Entropy (WCE) loss. The weight of each class $c$ is denoted as $w^c$, and is inversely proportional to the frequency of the class in the training and validation sets. The WCE loss is formulated as follows:
\begin{equation} \label{eq:loss_weighted_CE}
    \mathcal{L}_{\mbox{\footnotesize{task = classification}}}(\mathbf{P}, \mathbf{\hat{P}}) = - \sum_{i=1}^B \sum_{c=1}^C w^c p_i^c log(\hat{p}_i^c)
\end{equation}
where, $p_i^c$ and $\hat{p}_i^c$ denote the ground truth and the predicted probability of class $c$ , respectively, and
$\mathbf{P}$ and $\mathbf{\hat{P}}$ are the batched ground truth (one hot) and the predicted distributions.

\subsection{Missing values} \label{sec:missing_values}
It is not uncommon to have tabular data with missing values. Handling missing values is a crucial aspect of tabular data preprocessing, as it can significantly impact the quality of modeling and analysis. There exist different strategies to address this challenge. For example, the multi-modal generative approach proposed in~\citep{Dolci2022} which compensates for missing modalities via a three-module framework. In the first module, each modality is processed independently, the second one imputes the unavailable data using pretrained generators and the last module fuses all features prior to the prediction. 

In our study, we impute missing values using an iterative approach. The tabular data is presented as a matrix where each column represents an attribute and each row is a training set sample. In each step, one attribute (column) is selected as the output $z$, and the remaining columns are considered as inputs $X$. A regressor is trained on the non missing values of $(z,X)$,  and then employed to predict the missing values of $z$. This iterative process repeats for each of the columns. As in~\citep{wolf2022DAFT} we add a indicator, such as a NaN flag to indicate imputed values.

While this technique is widely used, it is important to acknowledge potential biases it may introduce. 
Often, the values of specific attributes are consistently absent from datasets which belong to specific classes. For instance, datasets related to CN subjects typically lack measurements that might pose health risks, requiring invasive procedures or imaging with ionizing radiation. In such cases, the presence of an indicator for missing values could itself be an 'unfair' cue, potentially leading to improved results for the 'wrong reasons. Additionally, the imputation of attributes with missing values in datasets related to one class might heavily rely on the existing values of another class, potentially compromising their validity.
Despite these concerns, we opted for the imputation strategy outlined above to ensure a fair comparison with existing methods that utilize it.

\subsection{Ensemble learning}  \label{sec:ensemble}
To enhance and stabilize the inference process, we use an ensemble model, denoted as $\mathcal{E}$, which aggregates the results of $M$ trained models, $\mathcal{F}_1$ to $\mathcal{F}_M$.
Addressing regression problems, the final prediction is a simple average of the predictions provided by all ensemble's models:
\begin{equation} \label{eq:ensemble_regression}
\mathcal{E}_{\mbox{\footnotesize{regression}}}(T, I) = \frac{1}{M} \sum_{m=1}^M \mathcal{F}_m(T, I).
\end{equation}

In multi-class classification tasks involving $C$ classes, each model $\mathcal{F}_m$ produces a probability distribution vector over the predicted classes, represented as $\mathbf{p}_m = (p^{c=1}, \ldots, p^{c=C})_m$. To obtain the final ensemble prediction, $\mathcal{E}_{classification}(T, I)$, these models are combined using the weighted average of their probability distributions, as follows:
\begin{equation} \label{eq:ensemble_output}
\begin{array}{l l}
    \displaystyle \mathcal{E}_{\mbox{\footnotesize{classification}}}(T, I) = \sum_{m=1}^M w_m\mathcal{F}_m(T,I) =
    \sum_{m=1}^M w_m \mathbf{p}_m.
\end{array}
\end{equation}
Here $w_m$ defines the weight of the prediction of $\mathcal{F}_m$ in the ensemble. 
In general, we wish $w_m$  to be higher for models with higher confidence (lower uncertainty). 
Assuming that the prediction uncertainty positively correlates with its entropy, we set $w_m$ to be inversely proportional to the entropy of network's prediction, as follows:
\begin{equation} \label{eq:weights4ensemble}
\begin{array}{l l}
    w_m & \displaystyle =  \frac{J(\mathcal{F}_m(T, I))}{\sum_{i=1}^M J(\mathcal{F}_i(T, I))}  =  \frac{J(\mathbf{p}_m)}{\sum_{i=1}^M J(\mathbf{p}_i)} ,
\end{array}
\end{equation}
where $J(\cdot)$ is defined by $\frac{1}{H}$ and $H$ is the entropy function. The term $\sum_{i=1}^M J(\mathbf{p}_i)$  serves as a normalization factor.

\subsection{Imaging-tabular data fusion applications}
To showcase the versatility of the proposed imaging-tabular data fusion framework, we explore two distinct medical imaging applications, each featuring a unique architectural configuration. These applications encompass brain age prediction conditioned by the subject's sex (Section~\ref{sec:method_brainage_using_hyper}) and multi-class classification of subjects into AD, MCI, and CN groups (Section~\ref{sec:method_AD_using_hyper}).

\subsubsection{Conditioned brain age prediction using hypernetworks}  \label{sec:method_brainage_using_hyper}
\begin{figure}[!t]
    \centering
    \includegraphics[scale=0.60]{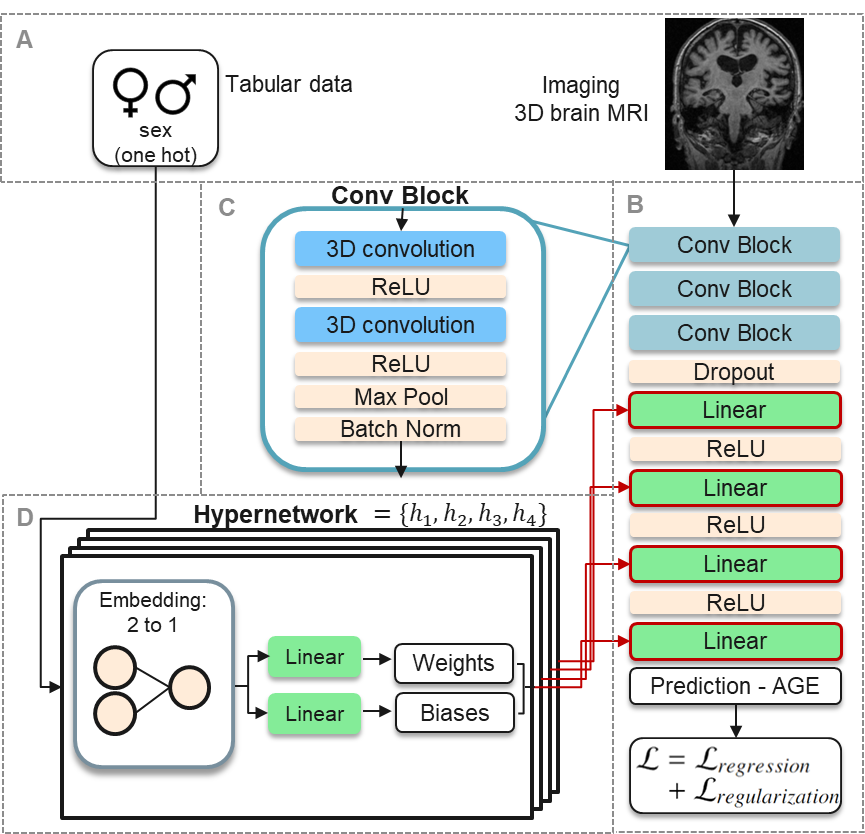}
    \caption{\textcolor{black}{HyperFusion architecture for conditioned brain age prediction.
    A: The inputs include the subjects' sex (encoded as a 2D one-hot vector) and the corresponding 3D brain MRIs. B: The primary network backbone is a {var}iant of the VGG architecture~\citep{simonyan2014VGG}, where the parameters of its final four linear layers (framed in red) are external and generated by the hypernetwork. C: A closer look of a convolutional block . D: The hypernetwork comprises four sub-networks ($h_1, h_2, h_3, h_4$), each corresponds to one of the linear layer in the primary network.
    }\label{fig:brainage_model}}
\end{figure}

The proposed hypernetwork framework for conditioned brain age prediction is illustrated in Figure~\ref{fig:brainage_model}. The primary network takes 3D brain MRI scans (T1w) of healthy subjects as input, while the tabular data (single attribute) fed into the hypernetwork consists of 2D one-hot vectors representing the subjects' sex.
\textcolor{black}{
The hypernetwork compound is trained to address a regression problem using the loss function defined in Equations~\ref{eq:loss_func} and~\ref{eq:loss_regression}. The primary network architecture (Figure~\ref{fig:brainage_model}B-C) is a variant of the VGG backbone~\citep{simonyan2014VGG}, as suggested in~\citep{levakov2020brainage} for brain age prediction. The parameters of its final four linear layers (framed in red) are external and generated by the hypernetwork. The network output the subjects' brain age ($\in \mathbb{R}$), and the loss function is a weighted sum  of the regression loss (MSE) and weight decay regularization, as defined in Equations~\ref{eq:loss_func}-\ref{eq:loss_regression}. Figure~\ref{fig:brainage_model}C provides a closer look of a convolutional block, comprising two convolutional layers with ReLU activation, batch normalization and downsampling through max-pooling. 
As illustrated in Figure~\ref{fig:brainage_model}D the hypernetwork comprises four sub-networks ($h_1, h_2, h_3, h_4$), each corresponds to one of the linear layer in the primary network. Each $h_k$ is composed of a 2-to-1 MLP for the embedding of the one-hot input vectors.
The embedding step is crucial, especially when using the orthogonal one-hot representation of the sex attribute (01 or 10). Without embedding, the hypernetwork would generate a distinct set of weights for each sex attribute value. We found that in this case, the most effective integration of the hypernetwork occurred within the linear layers at the final processing stage.} 

\subsubsection{AD classification using hypernetworks}  
\label{sec:method_AD_using_hyper}
\begin{figure}[!t]
    \centering
    \includegraphics[scale=0.60]{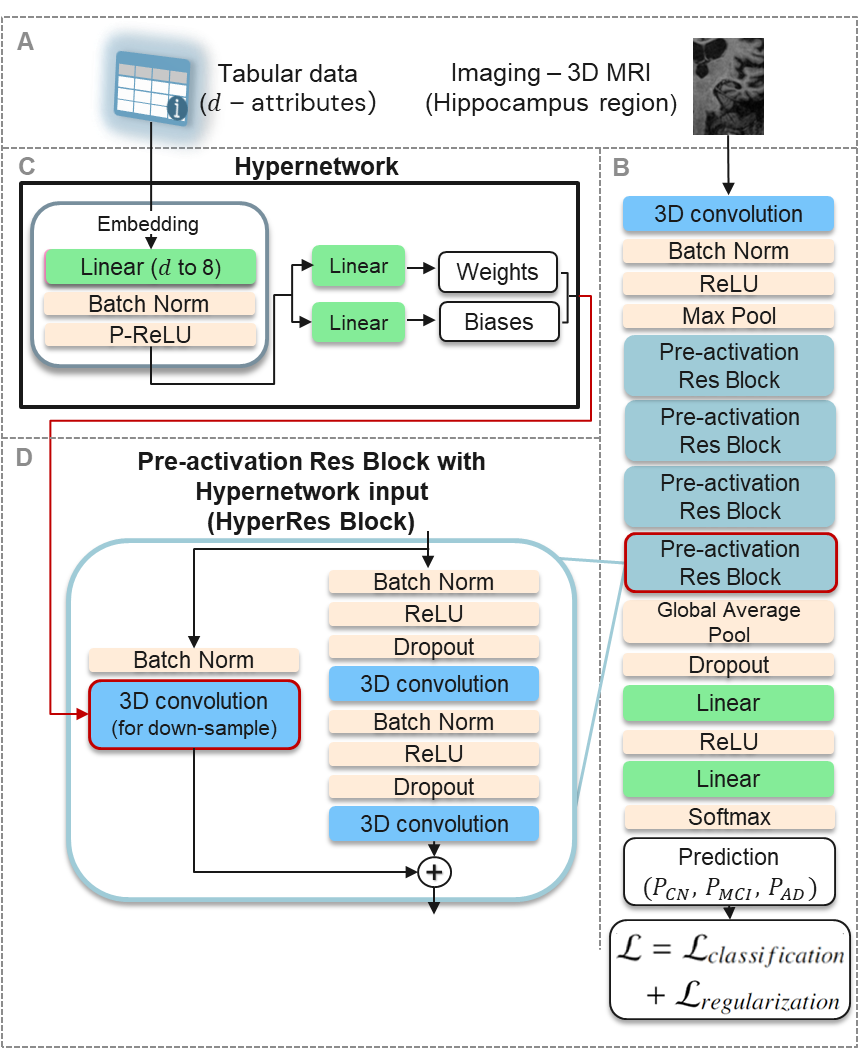}
    \caption{
    \textcolor{black}{\textbf{HyperFusion architecture for the AD classification}. A: The input consists of tabular attributes ($d$ in total) of the subjects along with their brain MRIs. B: The primary network's is composed of pre-activation ResNet blocks followed by two linear layers. The last ResNet block (framed in red) gets a subset of its parameters from the hypernetwork. 
    The primary network predictions are probability distributions ($P_{CN}, P_{MCI}, P_{AD}$) produced by the softmax layer.  The loss is a weighted sum of the classification loss (weighted CE) and weight decay regularization, as detailed in Equations~\ref{eq:loss_func},\ref{eq:loss_weighted_CE}. C: The hypernetwork architecture D: A closer look at the pre-activation Res Block. See text for details.
    }
    \label{fig:AD_model}}
\end{figure}

Figure~\ref{fig:AD_model} visually presents the hypernetwork framework for multi-class classification of subjects into CN, MCI, and AD groups. The input to the hypernetwork (Figure~ \ref{fig:AD_model}A) is the tabular data which comprise nine clinical and demographic attributes as described in Section~\ref{sec:exp_AD_tabular_data}. Similar to the brain age prediction's application, the primary network is fed with 3D brain MRI scans (Figure~ \ref{fig:AD_model}B). Yet, for AD classification, the images were cropped to include only the hippocampus and its surrounding tissues as suggested in~\citep{wen2020CNN_AD_overview}.  The cropping significantly reduced the input size leading eventually to a reduction in the number of the required network parameters and to better classification results. We note that the hippocampus is the significant component of the limbic lobe being a crucial region for learning and memory. Hippocampal atrophy is a known bio-marker for AD~\citep{Rao2022, Salta2023}.
The hypernetwork, illustrated in Figure~\ref{fig:AD_model}C is composed of a non-linear, single-layer MLP used for embedding, with Parametric ReLU (P-ReLU)~\citep{he2015PReLU}. The MLP generates weights and biases for the primary network.
The primary network predictions are the probability distributions ($P_{CN}, P_{MCI}, P_{AD}$) produced by the softmax layer. The loss is a weighted sum of the multi-class classification loss (WCE) and a weight decay regularization term, as detailed in Equations~\ref{eq:loss_func},\ref{eq:loss_weighted_CE}. 
The backbone of the primary network, depicted in Figure~\ref{fig:AD_model}B is based on the pre-activation ResNet blocks followed by two linear layers. The last ResNet block (framed in red), called the HyperRes block, gets a subset of its parameters from the hypernetwork. 
A closer view of a pre-activation ResNet blocks is presented in Figure~\ref{fig:AD_model}D. The architecture (based on~\citep{he2016preactivResnet}) is the same for all blocks, yet the illustration specifically refers to the HyperRes block which gets the parameters to one of its convolutional layer (framed in red) from the hypernetwork.

\section{Experiments} \label{sec:experiments}
In this section, we detail the experimental configurations, ablation studies, prediction outcomes, and comparisons for two different brain imaging analysis applications, aiming to showcase the robustness and adaptability of the proposed multi-modal fusion hypernetwork framework.
In particular, Section~\ref{sec:exp-BrainAgePrediction} delves into the experiments and findings related to brain age estimation
conditioned by the subject's sex, while Section~\ref{sec:exp-ADclassification} addresses multi-class classification of subjects into CN, MCI, or AD groups. \textcolor{black}{Additional experiments for binary AD classification are presented in~\ref{sec:AD_binary}.} 

\textcolor{black}{
\subsection{Experimental details}  \label{sec:exp-experimental_details}
All the experiments where conducted using a Tesla V100 32 GB GPU with PyTorch. \textcolor{black}{Table~\ref{tab:experimental_details} presents the experimental details of the AD classification and brain age prediction tasks.}
All architectural details of the networks are presented in Tables~\ref{tbl:BAModelArch},~\ref{tbl:ADModelArch} in~\ref{sec:Brainage_parameters} and~\ref{sec:AD_parameters}, respectively.}
 
\begin{table*}[!h]
\footnotesize
\caption{\label{tab:experimental_details}
\textcolor{black}{Experimental details of the AD classification and brain age prediction tasks}
}
\centering
\textcolor{black}{
\begin{tabular}{|l|c|c|c|c|c|c|c|}
\hline
\textbf{Experiment} & \textbf{epochs} & \textbf{batch size} & \textbf{optimizer} & \textbf{weight decay factor} & \textbf{learning rate} \\ \hline
Brain age prediction  & 70 & 64 & ADAM &  $.05$ & $1.5\cdot10^{-4}$  \\ \hline
AD classification  & 250 & 32 & ADAM &  $10^{-5}$ & $10^{-4}$  \\ \hline
\end{tabular}}
\end{table*}

\subsection{Brain age prediction conditioned by sex}  \label{sec:exp-BrainAgePrediction}
While most existing approaches address brain age estimation based on brain MRI data alone, we condition the brain age regression problem on the subject's sex. 
We first validate the hypothesis that sex information improves the accuracy of brain age prediction. We then evaluate the efficiency of the hypernetwork framework in fusing imaging and tabular data. This evaluation involves a comparison with the baseline model that exclusively relies on brain MRI data.

\subsubsection{The data}  \label{sec:brain_age_dataset}
The data for this study include 26,691 brain MRI scans of different healthy human subjects from 19 sources. 
The table in~\ref{sec:Brain-Age-dataset} presents a comprehensive information on each dataset including the number of subjects as well as their age and sex distributions.

\subsubsection{Training and evaluation}
We evaluate the hyperfusion and alternative brain-age prediction frameworks using the Mean Absolute Error (MAE) between the chronological and the predicted age. To ensure robustness, we trained multiple models with different random weight initialization to mitigate the potential impact of incidental results. The test results were calculated based on an ensemble of the five best-performing models identified during validation. The ensemble's output was derived through an average of the model predictions, as detailed in Section~\ref{sec:ensemble}, further enhancing the reliability of our findings.
\subsubsection{Preprocessing and partitioning}  \label{sec:preproc_brain_age}
We applied a preprocessing pipeline to the T1-weighted MRI scans as proposed in~\citep{levakov2020brainage} - see ~\ref{sec:Brain-Age-dataset} for details.
Before being utilized as model inputs, the images were standardized to attain a mean of zero and a standard deviation of one across all non-zero voxels, ensuring data uniformity for effective training.
We note that subjects' sex and age are routinely recorded during the scanning visits. Therefore, we can utilize almost all of the available scans without having to handle issues related to missing values.
The tabular data, in this case, consists of a single binary value which is the subject's sex - male or female. It is represented in out study by a 2D one-hot vector.
We partitioned the dataset into training (80\%), validation (10\%), and test (10\%) subsets, preserving consistent age and sex distributions across all the subsets.
\subsubsection{Differences in brain aging between males and females}  \label{sec:exp_brainage_check_hypothesis}

Prior to demonstrating the strength of the proposed hypernetwork integrating imaging and tabular data for the task of brain age prediction conditioned by the subject's sex we conducted a baseline experiment to support the assumption that the availability of sex information enhances the prediction results.   
We trained three CNNs, each tailored to a specific dataset: one dedicated to male brain scans, another to female brain scans, and a third to mixed data encompassing both sexes. The mixed dataset, comprising approximately half the volume of the complete dataset, ensured equitable sample representation for each model during training, facilitating a fair comparison. Our evaluations encompassed the entire test set, as well as two subsets—one exclusively featuring male subjects and the other exclusively featuring female subjects.

\begin{figure*}[!t]
    \centering
    \includegraphics[scale=0.80]{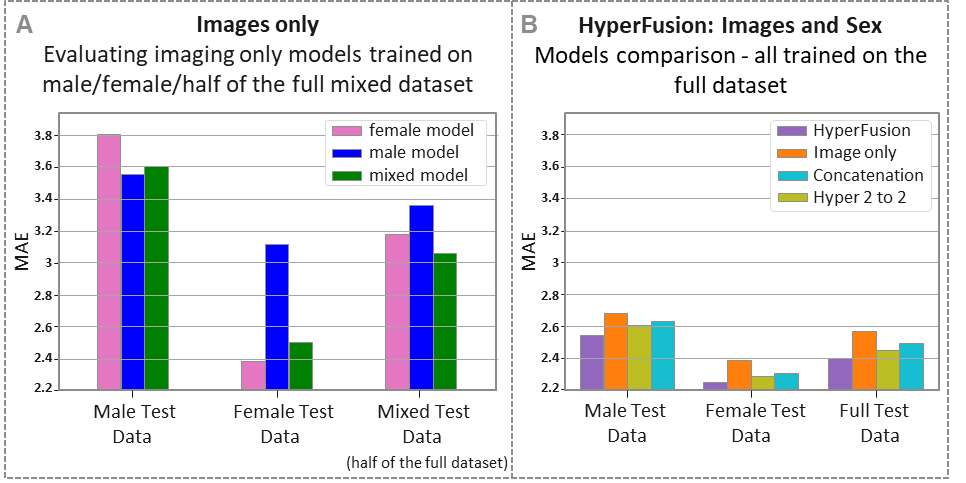}
    \caption{Conditioned brain age prediction results. A) An imaging only experiment. The MAE scores obtained for baseline networks trained on either male only (blue), female only (pink) and mixed (green) equally sized training sets, where the test sets are composed of male only (left), female only (middle) or mixed (right) subsets. \textcolor{black}{B) HyperFusion: A comparison between the baseline model using imaging data alone (orange), concatenation based fusion (light blue), Hyperfusion with different embedding - inputs 2d vector and outhputs 2d vector (olive green), and the proposed HyperFusion model (purple) using MAE metric. The inference was performed using either male test data (left), female test data (middle) or the entire (mixed) test dataset (right).}
    }
    \label{fig:brainage_results}
\end{figure*}

\textcolor{black}{The MAE scores of obtained by this set of experiments are presented in Figure \ref{fig:brainage_results}A. The blue, pink and green bars refer to network training with either male only, female only or mixed data, respectively -maintaining equal-size training sets.
The text sets are composed of male only (left), female only (middle) or mixed (right) subsets. The plot strengthens our 
hypothesis that sex information can facilitate brain age prediction - motivating the proposed hypernetwork framework.}

\subsubsection{\textcolor{black}{Comparisons and ablation study}} \label{sec:comparison_and_ablation}
\textcolor{black}{Figure~\ref{fig:brainage_results}B presents comparisons of the proposed hyperfusion approach (purple bars) with both image-only (orange bars) and common tabular-imaging concatenation (light blue bars) methods. The comparisons were made for male-only (left bars), female-only (middle bars) and the entire (right bars) test data. For a fair comparison with our hyperfusion model, which provides parameters to all four linear layers in the primary network, the concatenation model combined the sex and image features in each linear layer of the baseline network. Additionally, we performed an ablation study on the embedding of the hypernetwork, specifically comparing our proposed one-hot vector into a single scalar mapping (purple bars) with a mapping into a 2D vector (olive-colored bars). An ablation study assessing the proposed external layer selection can be found in~\ref{sec:hyperlayers-selection}.}
\subsubsection{\textcolor{black}{Results and discussion}}
\textcolor{black}{The comparisons presented in Figure~\ref{fig:brainage_results} clearly show that the proposed hyperfusion model outperforms the image-only network for all male-female train-test configurations, as well as the standard concatenation framework. Although beyond the scope of this paper, it is interesting to note that the prediction of female age is much more accurate than the prediction of male age (see the leftmost and middle bars in Figure~\ref{fig:brainage_results}A and B).
We assume that this might be due to the greater variability of brain age in males.}


\subsection{Multi-class AD classification} \label{sec:exp-ADclassification}
To further assess the efficiency of our network in fusing imaging and tabular data, we performed a set of experiments on the multi-class AD classification task conditioned by different number and compositions of tabular data attributes. 
We compared our results to those obtained using either of the data modalities as well as to other existing approaches. 

\subsubsection{The data}
\textcolor{black}{We used four phases of the ADNI databases (ADNI1, ADNI2, ADNI GO, and ADNI3) - selecting a single and complete dataset of each individual (obtained during the first visit). Each of the selected datasets includes a single T-1 MRI scan and the (temporally) corresponding EHR. Statistics of the ADNI data used are presented in~Table \ref{tab:AD_classification_data}. The information includes the number of samples "N" per label, the mean and standard deviation of age, and the male/female ratio within each diagnostic group. The data used along with the corresponding ADNI's subject identification numbers are available at the Git repository enclosed with this work.}

\begin{table}[!t]
\footnotesize
\caption{\label{tab:AD_classification_data}
Statistics on the CN, MCI and AD sub-populations in our data. 
 }
\centering
\begin{tabular}{|c|c|c|c|}
\hline

\textbf{Diagnosis} & \textbf{N (\%)} & \textbf{Age - mean (±std)} & \textbf{Sex (M:F)} \\ \hline
AD & 365 (17.2\%) & 75.1 (±7.8) & (198:167) \\ \hline
CN & 740 (34.9\%) & 72.2 (±6.8) & (309:431) \\ \hline
MCI & 1015 (47.9\%) & 72.8 (±7.6) & (569:446) \\ \hline
Over all & 2120 (100\%) & 73.0 (±7.4) & (1076:1044) \\ \hline

\end{tabular}
\end{table}

\subsubsection{Image preprocessing}
The preprocessing pipeline of the MRI scans is similar to the one in Section~\ref{sec:brain_age_dataset}. As discussed in Section~\ref{sec:method_AD_using_hyper}, we cropped the scans into two 3D sub-images of size $64\times 96 \times 64$ voxels, including either the left or the right hippocampus regions. 

\subsubsection{Tabular preprocessing} \label{sec:exp_AD_tabular_data}
While the subjects' EHRs contain numerous attributes we chose to use only nine of them.
The demographic attributes include age, sex, and education. The CSF biomarkers include A$\beta$42, P-tau181, and T-tau. The third attribute category encompasses composite measures derived from 18F-fluorodeoxyglucose (FDG) and florbetapir (AV45) PET scans. It is important to note that cognitive scores from the EHR were excluded from our study because they were directly used for AD diagnosis. In other words, these scores alone are sufficient for AD classification~\citep{Qiu2018ADfusion_cognitiveTests}, rendering the entire study meaningless.

Binary (sex) and other discrete attributes  were transformed into one-hot vectors. Conversely, attributes with continuous values were centralized and normalized to have zero mean and standard deviation of one. Missing values were handled as described in Section~\ref{sec:missing_values}.

\subsubsection{Data partitioning}  \label{sec:AD_data_partitioning}
The dataset is partitioned into five non-overlapping subsets of equal size, ensuring that each subset maintains a similar joint distribution of age, sex, and diagnosis, in accordance with the methodology outlined by~\citep{wen2020CNN_AD_overview}. Among these five subsets, four are allocated for cross-validation, while the fifth is set aside for testing, as depicted in~\ref{sec:AD_data_splitting}. The partitioning involves a random shuffling of the data using a predefined random seed, referred to as the split seed. This systematic approach to data splitting enhances the validity and generality of the evaluation process, showcasing the robustness of our model.

\subsubsection{Ablation study and comparisons} \label{sec:AD_ablation}
To evaluate the effectiveness of our proposed HyperFusion approach for image-tabular data fusion, we conducted a comprehensive analysis comparing it with numerous unimodal and multi-modal models. The unimodal models include a baseline MLP designed for tabular data and a pre-activation ResNet optimized for image data processing. Both the baseline MLP and the ResNet are components in our complete hypernetwork compound, as illustrated on the left-hand side of Figure~\ref{fig:AD_model}C and Figure~\ref{fig:AD_model}B, respectively.

The imaging-tabular fusion models we compared include various methods. The first one, dubbed `concatenation', involved employing our pre-activation ResNet backbone (presented in Figure \ref{fig:AD_model}B) yet, replacing the HyperRes block with a regular pre-activation Res block. The tabular attributes were concatenated with the second-to-last fully connected layer, a technique documented in prior works like~\citep{esmaeilzadeh2018concat_fusing}.

Additionally, we compared our method to the results reported for two contemporary imaging-tabular fusion techniques: the DAFT~\citep{wolf2022DAFT} and the late-fusion approach proposed by~\citep{prabhu2022multimodal_AD_entropydecision}.
To ensure a fair comparison and highlight the unique (isolated) contribution of our hypernetwork, we constructed and trained two additional networks. Specifically, we utilized our primary network backbone and training regimen, including our proposed regularized weighted categorical loss function, maintaining the methodologies of  FiLM~\citep{perez2018film} (FiLM-like implementation) and DAFT~\citep{wolf2022DAFT} (DAFT-like implementation). We note that our implementation of the FiLM was adapted from~\citep{wolf2022DAFT}, since originally the FiLM was designed for text and natural image fusion.

\begin{figure*}[!t]
    \centering
    \includegraphics[scale=0.79]{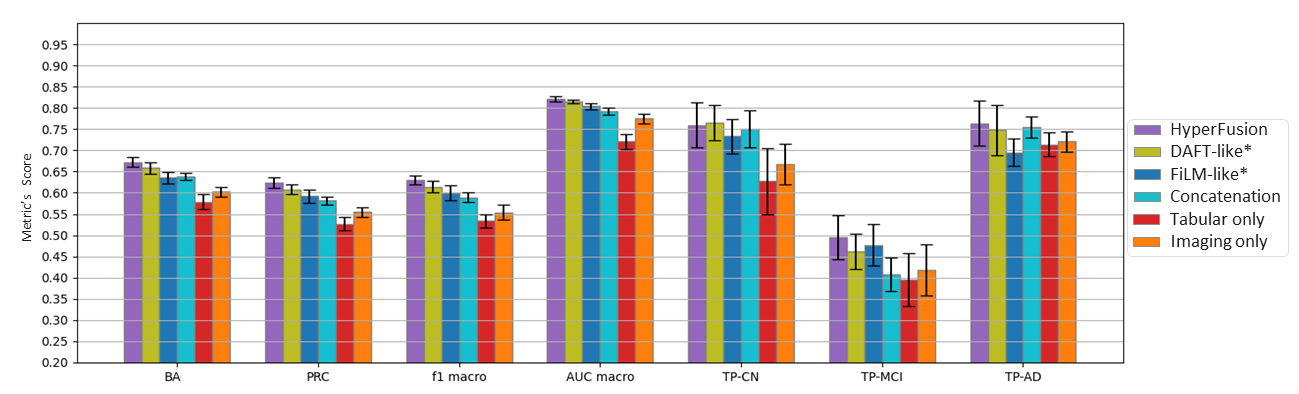}
    \caption{
    \textcolor{black}{Bar plot presentation of the AD classification results for six competing models and ours using seven different metrics (Section~\ref{sec:exp_AD_trainingNevaluation}). The compared methods including `DAFT-like$^{*}$' and `FiLM-like$^{*}$' are described in Section~\ref{sec:AD_ablation}.}
    }
    
    \label{fig:AD_bars_test_res}
\end{figure*}

\begin{table*}[!t]
\footnotesize
\caption{\label{tab:AD_results}
\textcolor{black}{AD classification results for the proposed method and other competing models using seven different metrics.}
}
\centering
\begin{tabular}{|l|c|c|c|c|c|c|c|}
\hline

\textbf{model} & \textbf{BA} & \textbf{PRC} & \textbf{f1 macro} & \textbf{AUC macro} & \textbf{TP-CN} & \textbf{TP-MCI} & \textbf{TP-AD} \\ \hline

Imaging only & 0.579 ± 0.018 & 0.527 ± 0.016 & 0.534 ± 0.016 & 0.721 ± 0.018 & 0.627 ± 0.078 & 0.395 ± 0.063 & 0.714 ± 0.028 \\
 & p $<$ 0.01 & p $<$ 0.01 & p $<$ 0.01 & p $<$ 0.01 & p $<$ 0.01 & p $<$ 0.01 & p $<$ 0.01 \\ \hline
Tabular only & 0.602 ± 0.011 & 0.555 ± 0.011 & 0.554 ± 0.018 & 0.775 ± 0.011 & 0.668 ± 0.048 & 0.418 ± 0.060 & 0.721 ± 0.024 \\
 & p $<$ 0.01 & p $<$ 0.01 & p $<$ 0.01 & p $<$ 0.01 & p $<$ 0.01 & p $<$ 0.01 & p $<$ 0.01 \\ \hline
Concatenation & 0.638 ± 0.008 & 0.582 ± 0.009 & 0.589 ± 0.011 & 0.792 ± 0.008 & 0.751 ± 0.044 & 0.408 ± 0.039 & 0.755 ± 0.025 \\
 & p $<$ 0.01 & p $<$ 0.01 & p $<$ 0.01 & p $<$ 0.01 & p = 0.550 & p $<$ 0.01 & p = 0.296 \\ \hline
FiLM-like implementation$^*$ & 0.635 ± 0.014 & 0.592 ± 0.016 & 0.600 ± 0.017 & 0.804 ± 0.007 & 0.733 ± 0.040 & 0.477 ± 0.049 & 0.696 ± 0.032 \\
 & p $<$ 0.01 & p $<$ 0.01 & p $<$ 0.01 & p $<$ 0.01 & p = 0.197 & p = 0.234 & p $<$ 0.01 \\ \hline
DAFT-like implementation$^*$ & 0.658 ± 0.014 & 0.608 ± 0.012 & 0.614 ± 0.014 & 0.815 ± 0.005 & 0.765 ± 0.041 & 0.462 ± 0.042 & 0.748 ± 0.060 \\
 & p $<$ 0.01 & p $<$ 0.01 & p $<$ 0.01 & p $<$ 0.01 & p = 0.717 & p = 0.038 & p = 0.224 \\ \hline
HyperFusion & 0.673 ± 0.012 & 0.624 ± 0.012 & 0.630 ± 0.011 & 0.822 ± 0.006 & 0.759 ± 0.053 & 0.495 ± 0.052 & 0.764 ± 0.052 \\ \hline

Reported DAFT results& 0.622 ± 0.044 & - & 0.600 ± 0.045 & - & 0.767 ± 0.080 &  0.449 ± 0.154 &  0.651 ± 0.144 \\ 
\citep{wolf2022DAFT}  & &&&&&&\\
\hline
Reported late-fusion results & 0.6330 & 0.6473 & 0.6240  & - & - &  - &  - \\ 
\citep{prabhu2022multimodal_AD_entropydecision}& &&&&&&\\
\hline
\multicolumn{6}{@{}l}{$^*$our adapted implementation as described in Section~\ref{sec:AD_ablation}}

\end{tabular}
\end{table*}
\subsubsection{Training and evaluation} \label{sec:exp_AD_trainingNevaluation}
We use the Adam optimizer for training, coupled with a regularized WCE loss (Equations \ref{eq:loss_func}, \ref{eq:loss_weighted_CE}), as outlined in Sections~\ref{sec:methods_loss} and~\ref{sec:method_AD_using_hyper}. The WCE loss accounts for the imbalance between the CN, MCI and AD classes in the training data. However, using standard categorical cross entropy loss while over-sampling the less frequent classes can handle the class imbalance problem as well. 
The embedding MLP, discussed in detail in Section~\ref{sec:method_AD_using_hyper} and depicted in Figure~\ref{fig:AD_model}C, underwent pre-training exclusively with the training tabular data to expedite the convergence of the entire network compound.

As mentioned in Section~\ref{sec:method_AD_using_hyper}, each subject's imaging data includes two 3D subimages of the left and right hippocampus. During training, we augment the dataset by randomly selecting either subimage in each feed-forward iteration. For better predictions during validation and testing, we processed each subject's data twice, once for each brain side, and averaged the soft decisions. Furthermore, we achieved enhanced test results employing an ensemble model, as elaborated in Section~\ref{sec:ensemble}.

To guarantee the robustness of our findings, we adopted an extensive cross-validation strategy encompassing three unique split seeds, each undergoing three rounds of random initialization (versions). In every cross-validation iteration, four models were trained and validated on distinct data folds. For effective result utilization, the four models from a specific version and split seed were aggregated to assess their collective performance on a shared, unseen test set. A visualization for better understanding the training, validation, and testing processes is provided in Figure~\ref{fig:data_split} in~\ref{sec:AD_data_splitting}.

We used various metrics to assess the performance of our framework and compare it with other models. These metrics include balanced accuracy (BA) and precision (PRC), defined as follows:
\begin{equation} \label{eq:BA}
\textcolor{black}{\mbox{BA} = \frac{1}{3}{\sum_c} \frac{TP_c}{TP_c+FN_c}} \end{equation}

\begin{equation} \label{eq:PRC}
\textcolor{black}{\mbox{PRC} = \frac{1}{3}{\sum_c} \frac{TP_c}{TP_c+FP_c}}
\end{equation}
Here, $c\in\{\mbox{CN, MCI, AD}\}$, and $TP_c$, $FN_c$, $FP_c$ represent the True Positive, False Negative, and False Positive values of class $c$, respectively. Additionally, we utilized the macro Area Under the ROC Curve (AUC) and macro F1 score. A macro AUC/F1 score is the arithmetic mean of all per-class AUC/F1 scores. These metrics consider the global performance of the models. For class-specific assessment of each model we also used the True Positive (TP) rate of each class (TP-CN, TP-MCI, TP-AD). Comprehensive confusion matrices are presented in~\ref{sec:AD_conf_matrix} (Figure~\ref{fig:conf_matrices}) for a thorough evaluation.

\subsubsection{Results and discussion}

In this section, we present a comprehensive comparison between all the models discussed in Section~\ref{sec:AD_ablation} and our hypernetwork model, evaluating both global and class-specific metrics. The results are summarized in Table~\ref{tab:AD_results} and Figure~\ref{fig:AD_bars_test_res}. 
In both the bar plot and the table we present the mean and standard deviation over the test data in all random splits and versions. The compared methods that include `DAFT-like$^{*}$' and `FiLM-like$^{*}$' are described in Section~\ref{sec:AD_ablation}.

It is noteworthy that all multi-modal methods consistently outperformed unimodal models in terms of global performance metrics and in nearly all class-specific metrics.

Table~\ref{tab:AD_results} also presents the scores reported for DAFT~\citep{wolf2022DAFT} and a late fusion method~\citep{prabhu2022multimodal_AD_entropydecision} for certain metrics, along with the results obtained for the methods we implemented as described in Section~\ref{sec:AD_ablation}. For a fair evaluation of the comparison, we note the following differences.

The late fusion method of~\citep{prabhu2022multimodal_AD_entropydecision} used datasets of 3,256 subjects (35.4\% CN, 51.1\% MCI, 13.5\% AD) from the same ADNI phases that we utilized. The tabular data contained an extensive set of 20 attributes, including patient demographics, medical history, vital signs, neurophysiological test results, cognitive function, and other relevant diagnostic information. Some of these attributes were deliberately excluded from our study, as noted in~\ref{sec:exp_AD_tabular_data}. Notably, the late fusion method obtained higher PRC scores at the expense of weaker BA results due to the inherent trade-off between these two metrics using imbalanced datasets. By applying class weight adjustments to the loss function during training, we were able to achieve better PRC and BA compared to the reported late fusion method results, despite using fewer subjects and a smaller set of tabular attributes (results are shown in~\ref{sec:AD_diff_class_weights}). The reported DAFT method~\citep{wolf2022DAFT} used datasets of 1,341 subjects (40.3\% CN, 40.1\% MCI, 19.6\% AD) from the same ADNI source and a similar set of tabular attributes. To address the difference in data size, we presented the scores obtained using our implementation of a DAFT-like method (Section~\ref{sec:AD_ablation}) as well.

Overall, the scores obtained for both global and class-specific metrics in comparison to other unimodal as well as multi-modal fusion methods demonstrate the superiority of the proposed hypernetwork framework. The p-values, calculated from the Mann-Whitney U tests, assess the statistical significance of the differences between the results.

\section{Discussion and Conclusions} \label{sec:discussion_conclusions}

We introduced hypernetworks as a solution to the intricate task of integrating tabular and imaging data. Additionally, we illustrated how these networks can condition the analysis of patients' brain MRIs based on their EHR data. This interaction between imaging and tabular information enhances the understanding of medical data, facilitating a comprehensive diagnostic process.

The robustness and generality of our method were validated through two distinct clinical applications. The core concepts proposed are adaptable to different network architectures. Furthermore, the straightforward integration of pre-trained networks into our framework promises accelerated convergence and heightened accuracy. Altogether, these aspects position our method as a practical and versatile tool for data fusion.

Our hypernetwork approach surpasses models relying solely on either imaging or tabular data. Unlike many fusion methods, which often use simple concatenation or mixed decisions based on both modalities, including some advanced ones that limit themselves to specific transformations, our method stands out for its increased versatility and degrees of freedom, ensuring superior data utilization in our architectures. Remarkably, when benchmarked against state-of-the-art fusion techniques, our approach not only delivers superior results but also demonstrates their statistical significance.

In this study, we condition the processing of images on the patient's EHR, adopting an intuitive approach reminiscent of viewing the image through the lens of the patient's tabular data. An intriguing avenue for future research involves reversing this process: conditioning the EHR analysis on the image. Furthermore, beyond the realms of CNNs and MLPs, exploring the integration of hypernetworks with other architectures opens up a promising pathway. Such adaptations could pave the way for robust data fusion techniques across a broader spectrum of data modalities.

In summary, this research not only identifies and tackles a crucial deep-learning challenge but also provides a robust solution through hypernetworks. The potential implications of this work extend from improving diagnostics to shaping the broader landscape of personalized medical care. As DNNs in the medical domain continue their rapid evolution, methods like ours, which bridge the gap between diverse data modalities, are poised to play a significant role.

\section*{Acknowledgments}

\textbf{Funding}:
The paper was partially supported by the Ministry of Health ( MoH 3-18509. T.R.R.) and the Israel Science Foundation (ISF 2497/19 T.R.R.)
Partial funding was provided by the Natural Sciences and Engineering Research Council of Canada (RGPIN-2023-05152), the Canadian Institute for Advanced Research (CIFAR) Artificial Intelligence Chairs program, the Mila - Quebec AI Institute technology transfer program, Calcul Quebec, and the Digital Research Alliance of Canada (alliance.can.ca). 

\textbf{Datasets}:
Uk-Biobank - This research has been conducted using the UK Biobank Resource (www.ukbiobank.ac.uk).

CORR - Data were provided in part by the Consortium for Reliability and Reproducibility (http://fcon\_1000.projects.nitrc.org/indi/CoRR/html/index.html)

ADNI - Data collection and sharing for this project was funded by the Alzheimer's Disease Neuroimaging Initiative (ADNI) (National Institutes of Health Grant U01 AG024904) and DOD ADNI (Department of Defense award number W81XWH-12-2-0012). ADNI is funded by the National Institute on Aging, the National Institute of Biomedical Imaging and Bioengineering, and through generous contributions from the following: AbbVie, Alzheimer's Association; Alzheimer's Drug Discovery Foundation; Araclon Biotech; BioClinica, Inc.; Biogen; Bristol-Myers Squibb Company; CereSpir, Inc.; Cogstate; Eisai Inc.; Elan Pharmaceuticals, Inc.; Eli Lilly and Company; EuroImmun; F. Hoffmann-La Roche Ltd and its affiliated company Genentech, Inc.; Fujirebio; GE Healthcare; IXICO Ltd.;Janssen Alzheimer Immunotherapy Research \& Development, LLC.; Johnson \& Johnson Pharmaceutical Research \& Development LLC.; Lumosity; Lundbeck; Merck \& Co., Inc.;Meso Scale Diagnostics, LLC.; NeuroRx Research; Neurotrack Technologies; Novartis Pharmaceuticals Corporation; Pfizer Inc.; Piramal Imaging; Servier; Takeda Pharmaceutical Company; and Transition Therapeutics. The Canadian Institutes of Health Research is providing funds to support ADNI clinical sites in Canada. Private sector contributions are facilitated by the Foundation for the National Institutes of Health (www.fnih.org). The grantee organization is the Northern California Institute for Research and Education, and the study is coordinated by the Alzheimer's Therapeutic Research Institute at the University of Southern California. ADNI data are disseminated by the Laboratory for Neuro Imaging at the University of Southern California.

GSP - Datawere provided in part by the Brain Genomics Superstruct Project of Harvard University and the Massachusetts General Hospital, (Principal Investigators: Randy Buckner, Joshua Roffman, and Jordan Smoller),with support from the Center for Brain Science Neuroinformatics Research Group, the Athinoula A. Martinos Center for Biomedical Imaging, and the Center for Human Genetic Research. Twenty individual investigators at Harvard and MGH generously contributed data toGSP.

FCP - Data were provided in part by the Functional Connectomes   Project   (https://www.nitrc.org/projects/fcon\_1000/).

ABIDE - Primary support for the work by Adriana Di Martino, and Michael P. Milham and his team was provided by the NIMH (K23MH087770), the Leon Levy Foundation, Joseph P. Healy and the Stavros Niarchos Foundation to the Child Mind Institute, NIMH award to MPM (R03MH096321), National Institute of Mental Health (NIMH5R21MH107045), Nathan S. Kline Institute of Psychiatric Research), Phyllis Green and Randolph Cowen to the Child Mind Institute.

PPMI - Data used in the preparation of this article were obtained from the Parkinson's Progression Markers Initiative (PPMI) database (www.ppmi-info.org/data). For up-to-date information on the study, visit www.ppmi-info.org.  PPMI—a public-private partnership—is funded by the Michael J. Fox Foundation for Parkinson's Research and funding partners, including [list of the full names of all of the PPMI funding partners can be found at www.ppmi-info.org/fundingpartners].

ICBM - Data used in the preparation of this work were obtained from the International Consortium for Brain Mapping (ICBM)  database  (www.loni.usc.edu/ICBM).  The  ICBM  project (Principal Investigator John Mazziotta, M.D., University of California, Los Angeles) is supported by the National Institute of Biomedical Imaging and BioEngineering. ICBM is the result of efforts of co-investigators from UCLA, Montreal Neurologic Institute, University of Texas at San Antonio, and the Institute of Medicine, Juelich/ Heinrich Heine University-Germany.

AIBL - Data used in the preparation of this article was obtained from the Australian Imaging Bio-markers and Lifestyle flagship study of aging (AIBL) funded by the Commonwealth Scientific and Industrial Research Organization (CSIRO) which was made available at the ADNI database (www.loni.usc.edu/ADNI). The AIBL researchers contributed data but did not participate in analysis or writing of this report. AIBL researchers are listed at www.aibl.csiro.au.

SLIM - Data were provided by the South-west University Longitudinal Imaging Multimodal (SLIM) Brain Data Repository (http://fcon\_1000.projects.nitrc.org/indi/retro/southwestuni\_qiu \_index.html).

IXI - Data were provided in part by the IXI database (http://brain-development.org/).

OASIS - OASIS is made available by Dr. Randy Buckner at the Howard Hughes Medical Institute (HHMI) at Harvard University, the Neuroinformatics Research Group (NRG) at Washington University School of Medicine, and the Biomedical Informatics Research Network (BIRN). Support for the acquisition of this data and for data analysis was provided by NIH grants P50 AG05681, P01 AG03991, P20MH071616, RR14075, RR 16594, U24 RR21382, the Alzheimer's Association, the James S. McDonnell Foundation, the Mental Illness and Neuroscience Discovery Institute, and HHMI.

CNP - This work was supported by the Consortium for Neuropsychiatric Phenomics (NIH Roadmap for Medical Research grants UL1-DE019580, RL1MH083268, RL1MH083269, RL1DA024853, RL1MH083270, RL1LM009833, PL1MH083271, and PL1NS062410).

COBRE - Data were provided by the Center for Biomedical Research Excellence (COBRE) (http://fcon\_1000.projects.nitrc.org/indi/retro/cobre.html).

CANDI - Data were provided in part by the Child and Adolescent Neuro Development Initiative - Schizophrenia Bulletin 2008 project.

Brainomics - Data were provided in part by the Brainomics project (http://brainomics.cea.fr/).

CamCAN - Data collection and sharing for this project was provided by the Cambridge Centre for Ageing and Neuroscience (CamCAN). CamCAN funding was provided by the UK Biotechnology and Biological Sciences Research Council (grant number BB/H008217/1), together with support from the UK Medical Research Council and University of Cambridge, UK.

\bibliographystyle{model2-names.bst}\biboptions{authoryear}
\bibliography{references, ref_datasets}

\begin{thebibliography}{73}
\expandafter\ifx\csname natexlab\endcsname\relax\def\natexlab#1{#1}\fi
\providecommand{\url}[1]{\texttt{#1}}
\providecommand{\href}[2]{#2}
\providecommand{\path}[1]{#1}
\providecommand{\DOIprefix}{doi:}
\providecommand{\ArXivprefix}{arXiv:}
\providecommand{\URLprefix}{URL: }
\providecommand{\Pubmedprefix}{pmid:}
\providecommand{\doi}[1]{\href{http://dx.doi.org/#1}{\path{#1}}}
\providecommand{\Pubmed}[1]{\href{pmid:#1}{\path{#1}}}
\providecommand{\bibinfo}[2]{#2}
\ifx\xfnm\relax \def\xfnm[#1]{\unskip,\space#1}\fi
\bibitem[{Aharon and Ben-Artzi(2023)}]{aharon2023}
\bibinfo{author}{Aharon, S.}, \bibinfo{author}{Ben-Artzi, G.}, \bibinfo{year}{2023}.
\newblock \bibinfo{title}{Hypernetwork-based adaptive image restoration}, in: \bibinfo{booktitle}{ICASSP 2023-2023 IEEE International Conference on Acoustics, Speech and Signal Processing (ICASSP)}, \bibinfo{organization}{IEEE}. pp. \bibinfo{pages}{1--5}.
\bibitem[{Alexander et~al.(2017)Alexander, Escalera, Ai, Andreotti, Febre, Mangone, Vega-Potler, Langer, Alexander, Kovacs et~al.}]{alexander2017HBN}
\bibinfo{author}{Alexander, L.M.}, \bibinfo{author}{Escalera, J.}, \bibinfo{author}{Ai, L.}, \bibinfo{author}{Andreotti, C.}, \bibinfo{author}{Febre, K.}, \bibinfo{author}{Mangone, A.}, \bibinfo{author}{Vega-Potler, N.}, \bibinfo{author}{Langer, N.}, \bibinfo{author}{Alexander, A.}, \bibinfo{author}{Kovacs, M.}, et~al., \bibinfo{year}{2017}.
\newblock \bibinfo{title}{An open resource for transdiagnostic research in pediatric mental health and learning disorders}.
\newblock \bibinfo{journal}{Scientific data} \bibinfo{volume}{4}, \bibinfo{pages}{1--26}.
\bibitem[{B{\"a}ckstr{\"o}m et~al.(2018)B{\"a}ckstr{\"o}m, Nazari, Gu and Jakola}]{backstrom2018CN_ADclass}
\bibinfo{author}{B{\"a}ckstr{\"o}m, K.}, \bibinfo{author}{Nazari, M.}, \bibinfo{author}{Gu, I.Y.H.}, \bibinfo{author}{Jakola, A.S.}, \bibinfo{year}{2018}.
\newblock \bibinfo{title}{An efficient {3D} deep convolutional network for {A}lzheimer's disease diagnosis using {MR} images}, in: \bibinfo{booktitle}{2018 IEEE 15th International Symposium on Biomedical Imaging (ISBI 2018)}, pp. \bibinfo{pages}{149--153}.
\bibitem[{Biswal et~al.(2010)Biswal, Mennes, Zuo, Gohel, Kelly, Smith, Beckmann, Adelstein, Buckner, Colcombe et~al.}]{biswal2010FCP}
\bibinfo{author}{Biswal, B.B.}, \bibinfo{author}{Mennes, M.}, \bibinfo{author}{Zuo, X.N.}, \bibinfo{author}{Gohel, S.}, \bibinfo{author}{Kelly, C.}, \bibinfo{author}{Smith, S.M.}, \bibinfo{author}{Beckmann, C.F.}, \bibinfo{author}{Adelstein, J.S.}, \bibinfo{author}{Buckner, R.L.}, \bibinfo{author}{Colcombe, S.}, et~al., \bibinfo{year}{2010}.
\newblock \bibinfo{title}{Toward discovery science of human brain function}.
\newblock \bibinfo{journal}{Proceedings of the national academy of sciences} \bibinfo{volume}{107}, \bibinfo{pages}{4734--4739}.
\bibitem[{Borisov et~al.(2022)Borisov, Leemann, Se{\ss}ler, Haug, Pawelczyk and Kasneci}]{borisov2022surveyDNN_tabular}
\bibinfo{author}{Borisov, V.}, \bibinfo{author}{Leemann, T.}, \bibinfo{author}{Se{\ss}ler, K.}, \bibinfo{author}{Haug, J.}, \bibinfo{author}{Pawelczyk, M.}, \bibinfo{author}{Kasneci, G.}, \bibinfo{year}{2022}.
\newblock \bibinfo{title}{Deep neural networks and tabular data: a survey}.
\newblock \bibinfo{journal}{IEEE Transactions on Neural Networks and Learning Systems} .
\bibitem[{Buckner et~al.(2014)Buckner, Roffman and Smoller}]{DVN25833_2014GSP}
\bibinfo{author}{Buckner, R.L.}, \bibinfo{author}{Roffman, J.L.}, \bibinfo{author}{Smoller, J.W.}, \bibinfo{year}{2014}.
\newblock \bibinfo{title}{{Brain Genomics Superstruct Project (GSP)}}.
\bibitem[{Carass et~al.(2017)Carass, Roy, Jog et~al.}]{carass2017longitudinal}
\bibinfo{author}{Carass, A.}, \bibinfo{author}{Roy, S.}, \bibinfo{author}{Jog, A.}, et~al., \bibinfo{year}{2017}.
\newblock \bibinfo{title}{Longitudinal multiple sclerosis lesion segmentation: resource and challenge}.
\newblock \bibinfo{journal}{NeuroImage} \bibinfo{volume}{148}, \bibinfo{pages}{77--102}.
\bibitem[{Chang et~al.(2019)Chang, Flokas and Lipson}]{chang2019hyper_weights_init}
\bibinfo{author}{Chang, O.}, \bibinfo{author}{Flokas, L.}, \bibinfo{author}{Lipson, H.}, \bibinfo{year}{2019}.
\newblock \bibinfo{title}{Principled weight initialization for hypernetworks}, in: \bibinfo{booktitle}{International Conference on Learning Representations}.
\bibitem[{Chieregato et~al.(2022)Chieregato, Frangiamore, Morassi, Baresi, Nici, Bassetti, Bn{\`a} and Galelli}]{chieregato2022COVID_CTandEHR}
\bibinfo{author}{Chieregato, M.}, \bibinfo{author}{Frangiamore, F.}, \bibinfo{author}{Morassi, M.}, \bibinfo{author}{Baresi, C.}, \bibinfo{author}{Nici, S.}, \bibinfo{author}{Bassetti, C.}, \bibinfo{author}{Bn{\`a}, C.}, \bibinfo{author}{Galelli, M.}, \bibinfo{year}{2022}.
\newblock \bibinfo{title}{A hybrid machine learning/deep learning {COVID}-19 severity predictive model from {CT} images and clinical data}.
\newblock \bibinfo{journal}{Scientific reports} \bibinfo{volume}{12}, \bibinfo{pages}{4329}.
\bibitem[{Coffey et~al.(1998)Coffey, Lucke, Saxton, Ratcliff, Unitas, Billig and Bryan}]{coffey1998AgeAndSex}
\bibinfo{author}{Coffey, C.E.}, \bibinfo{author}{Lucke, J.F.}, \bibinfo{author}{Saxton, J.A.}, \bibinfo{author}{Ratcliff, G.}, \bibinfo{author}{Unitas, L.J.}, \bibinfo{author}{Billig, B.}, \bibinfo{author}{Bryan, R.N.}, \bibinfo{year}{1998}.
\newblock \bibinfo{title}{Sex differences in brain aging: a quantitative magnetic resonance imaging study}.
\newblock \bibinfo{journal}{Archives of neurology} \bibinfo{volume}{55}, \bibinfo{pages}{169--179}.
\bibitem[{Cole and Franke(2017)}]{cole2017brainage_biomarker}
\bibinfo{author}{Cole, J.H.}, \bibinfo{author}{Franke, K.}, \bibinfo{year}{2017}.
\newblock \bibinfo{title}{Predicting age using neuroimaging: innovative brain ageing biomarkers}.
\newblock \bibinfo{journal}{Trends in neurosciences} \bibinfo{volume}{40}, \bibinfo{pages}{681--690}.
\bibitem[{Di~Martino et~al.(2014)Di~Martino, Yan, Li, Denio, Castellanos, Alaerts, Anderson, Assaf, Bookheimer, Dapretto et~al.}]{di2014ABIDE}
\bibinfo{author}{Di~Martino, A.}, \bibinfo{author}{Yan, C.G.}, \bibinfo{author}{Li, Q.}, \bibinfo{author}{Denio, E.}, \bibinfo{author}{Castellanos, F.X.}, \bibinfo{author}{Alaerts, K.}, \bibinfo{author}{Anderson, J.S.}, \bibinfo{author}{Assaf, M.}, \bibinfo{author}{Bookheimer, S.Y.}, \bibinfo{author}{Dapretto, M.}, et~al., \bibinfo{year}{2014}.
\newblock \bibinfo{title}{The autism brain imaging data exchange: towards a large-scale evaluation of the intrinsic brain architecture in autism}.
\newblock \bibinfo{journal}{Molecular psychiatry} \bibinfo{volume}{19}, \bibinfo{pages}{659--667}.
\bibitem[{Dolci et~al.(2022)Dolci, Rahaman, Chen, Duan, Fu, Abrol, Menegaz and Calhoun}]{Dolci2022}
\bibinfo{author}{Dolci, G.}, \bibinfo{author}{Rahaman, M.A.}, \bibinfo{author}{Chen, J.}, \bibinfo{author}{Duan, K.}, \bibinfo{author}{Fu, Z.}, \bibinfo{author}{Abrol, A.}, \bibinfo{author}{Menegaz, G.}, \bibinfo{author}{Calhoun, V.D.}, \bibinfo{year}{2022}.
\newblock \bibinfo{title}{A deep generative multimodal imaging genomics framework for alzheimer's disease prediction}, in: \bibinfo{booktitle}{2022 IEEE 22nd International Conference on Bioinformatics and Bioengineering (BIBE)}, pp. \bibinfo{pages}{41--44}.
\bibitem[{El-Sappagh et~al.(2020)El-Sappagh, Abuhmed, {Riazul Islam} and Kwak}]{el_appagh2020Multimodalmultitask}
\bibinfo{author}{El-Sappagh, S.}, \bibinfo{author}{Abuhmed, T.}, \bibinfo{author}{{Riazul Islam}, S.}, \bibinfo{author}{Kwak, K.S.}, \bibinfo{year}{2020}.
\newblock \bibinfo{title}{Multimodal multitask deep learning model for {A}lzheimer’s disease progression detection based on time series data}.
\newblock \bibinfo{journal}{Neurocomputing} \bibinfo{volume}{412}, \bibinfo{pages}{197--215}.
\bibitem[{Ellis et~al.(2009)Ellis, Bush, Darby, De~Fazio, Foster, Hudson, Lautenschlager, Lenzo, Martins, Maruff et~al.}]{ellis2009AIBL}
\bibinfo{author}{Ellis, K.A.}, \bibinfo{author}{Bush, A.I.}, \bibinfo{author}{Darby, D.}, \bibinfo{author}{De~Fazio, D.}, \bibinfo{author}{Foster, J.}, \bibinfo{author}{Hudson, P.}, \bibinfo{author}{Lautenschlager, N.T.}, \bibinfo{author}{Lenzo, N.}, \bibinfo{author}{Martins, R.N.}, \bibinfo{author}{Maruff, P.}, et~al., \bibinfo{year}{2009}.
\newblock \bibinfo{title}{The australian imaging, biomarkers and lifestyle (aibl) study of aging: methodology and baseline characteristics of 1112 individuals recruited for a longitudinal study of alzheimer's disease}.
\newblock \bibinfo{journal}{International psychogeriatrics} \bibinfo{volume}{21}, \bibinfo{pages}{672--687}.
\bibitem[{Esmaeilzadeh et~al.(2018)Esmaeilzadeh, Belivanis, Pohl and Adeli}]{esmaeilzadeh2018concat_fusing}
\bibinfo{author}{Esmaeilzadeh, S.}, \bibinfo{author}{Belivanis, D.I.}, \bibinfo{author}{Pohl, K.M.}, \bibinfo{author}{Adeli, E.}, \bibinfo{year}{2018}.
\newblock \bibinfo{title}{End-to-end {A}lzheimer's disease diagnosis and biomarker identification}, in: \bibinfo{editor}{Shi, Y.}, \bibinfo{editor}{Suk, H.I.}, \bibinfo{editor}{Liu, M.} (Eds.), \bibinfo{booktitle}{Machine Learning in Medical Imaging}, \bibinfo{publisher}{Springer International Publishing}, \bibinfo{address}{Cham}. pp. \bibinfo{pages}{337--345}.
\bibitem[{Feng et~al.(2020)Feng, Lipton, Yang, Small, Provenzano, Initiative, Initiative et~al.}]{feng2020BrainAgeDNN}
\bibinfo{author}{Feng, X.}, \bibinfo{author}{Lipton, Z.C.}, \bibinfo{author}{Yang, J.}, \bibinfo{author}{Small, S.A.}, \bibinfo{author}{Provenzano, F.A.}, \bibinfo{author}{Initiative, A.D.N.}, \bibinfo{author}{Initiative, F.L.D.N.}, et~al., \bibinfo{year}{2020}.
\newblock \bibinfo{title}{Estimating brain age based on a uniform healthy population with deep learning and structural magnetic resonance imaging}.
\newblock \bibinfo{journal}{Neurobiology of aging} \bibinfo{volume}{91}, \bibinfo{pages}{15--25}.
\bibitem[{Franke and Gaser(2019)}]{franke2019BrainAGE10years}
\bibinfo{author}{Franke, K.}, \bibinfo{author}{Gaser, C.}, \bibinfo{year}{2019}.
\newblock \bibinfo{title}{Ten years of {B}rain{AGE} as a neuroimaging biomarker of brain aging: what insights have we gained?}
\newblock \bibinfo{journal}{Frontiers in neurology} , \bibinfo{pages}{789}.
\bibitem[{Fratiglioni et~al.(1991)Fratiglioni, Grut, Forsell, Viitanen, Grafstr{\"o}m, Holmen, Ericsson, B{\"a}ckman, Ahlbom and Winblad}]{Fratiglioni1886AD_and_education}
\bibinfo{author}{Fratiglioni, L.}, \bibinfo{author}{Grut, M.}, \bibinfo{author}{Forsell, Y.}, \bibinfo{author}{Viitanen, M.}, \bibinfo{author}{Grafstr{\"o}m, M.}, \bibinfo{author}{Holmen, K.}, \bibinfo{author}{Ericsson, K.}, \bibinfo{author}{B{\"a}ckman, L.}, \bibinfo{author}{Ahlbom, A.}, \bibinfo{author}{Winblad, B.}, \bibinfo{year}{1991}.
\newblock \bibinfo{title}{Prevalence of {A}lzheimer{\textquoteright}s disease and other dementias in an elderly urban population}.
\newblock \bibinfo{journal}{Neurology} \bibinfo{volume}{41}, \bibinfo{pages}{1886--1886}.
\bibitem[{Frazier et~al.(2008)Frazier, Hodge, Breeze, Giuliano, Terry, Moore, Kennedy, Lopez-Larson, Caviness, Seidman et~al.}]{frazier2008CANDI}
\bibinfo{author}{Frazier, J.A.}, \bibinfo{author}{Hodge, S.M.}, \bibinfo{author}{Breeze, J.L.}, \bibinfo{author}{Giuliano, A.J.}, \bibinfo{author}{Terry, J.E.}, \bibinfo{author}{Moore, C.M.}, \bibinfo{author}{Kennedy, D.N.}, \bibinfo{author}{Lopez-Larson, M.P.}, \bibinfo{author}{Caviness, V.S.}, \bibinfo{author}{Seidman, L.J.}, et~al., \bibinfo{year}{2008}.
\newblock \bibinfo{title}{Diagnostic and sex effects on limbic volumes in early-onset bipolar disorder and schizophrenia}.
\newblock \bibinfo{journal}{Schizophrenia bulletin} \bibinfo{volume}{34}, \bibinfo{pages}{37--46}.
\bibitem[{Glorot and Bengio(2010)}]{glorot2010xavier_init}
\bibinfo{author}{Glorot, X.}, \bibinfo{author}{Bengio, Y.}, \bibinfo{year}{2010}.
\newblock \bibinfo{title}{Understanding the difficulty of training deep feedforward neural networks}, in: \bibinfo{booktitle}{Proceedings of the thirteenth international conference on artificial intelligence and statistics}, \bibinfo{organization}{JMLR Workshop and Conference Proceedings}. pp. \bibinfo{pages}{249--256}.
\bibitem[{Gorgolewski et~al.(2011)Gorgolewski, Burns, Madison, Clark, Halchenko, Waskom and Ghosh}]{gorgolewski2011nipype}
\bibinfo{author}{Gorgolewski, K.}, \bibinfo{author}{Burns, C.D.}, \bibinfo{author}{Madison, C.}, \bibinfo{author}{Clark, D.}, \bibinfo{author}{Halchenko, Y.O.}, \bibinfo{author}{Waskom, M.L.}, \bibinfo{author}{Ghosh, S.S.}, \bibinfo{year}{2011}.
\newblock \bibinfo{title}{Nipype: a flexible, lightweight and extensible neuroimaging data processing framework in python}.
\newblock \bibinfo{journal}{Frontiers in neuroinformatics} \bibinfo{volume}{5}, \bibinfo{pages}{13}.
\bibitem[{Ha et~al.(2016)Ha, Dai and Quoc}]{ha2016hypernetworks}
\bibinfo{author}{Ha, D.}, \bibinfo{author}{Dai, A.M.}, \bibinfo{author}{Quoc, V.L.}, \bibinfo{year}{2016}.
\newblock \bibinfo{title}{Hypernetworks. {C}o{RR} abs/1609.09106 (2016)}.
\newblock \bibinfo{journal}{arXiv preprint arXiv:1609.09106} .
\bibitem[{Hager et~al.(2023)Hager, Menten and Rueckert}]{Hager2023contrastive_cvpr}
\bibinfo{author}{Hager, P.}, \bibinfo{author}{Menten, M.J.}, \bibinfo{author}{Rueckert, D.}, \bibinfo{year}{2023}.
\newblock \bibinfo{title}{Best of both worlds: Multimodal contrastive learning with tabular and imaging data}, in: \bibinfo{booktitle}{Proceedings of the IEEE/CVF Conference on Computer Vision and Pattern Recognition (CVPR)}, pp. \bibinfo{pages}{23924--23935}.
\bibitem[{He et~al.(2015a)He, Zhang, Ren and Sun}]{he2015Kaiming_init}
\bibinfo{author}{He, K.}, \bibinfo{author}{Zhang, X.}, \bibinfo{author}{Ren, S.}, \bibinfo{author}{Sun, J.}, \bibinfo{year}{2015}a.
\newblock \bibinfo{title}{Delving deep into rectifiers: Surpassing human-level performance on imagenet classification}, in: \bibinfo{booktitle}{Proceedings of the IEEE international conference on computer vision}, pp. \bibinfo{pages}{1026--1034}.
\bibitem[{He et~al.(2015b)He, Zhang, Ren and Sun}]{he2015PReLU}
\bibinfo{author}{He, K.}, \bibinfo{author}{Zhang, X.}, \bibinfo{author}{Ren, S.}, \bibinfo{author}{Sun, J.}, \bibinfo{year}{2015}b.
\newblock \bibinfo{title}{Delving deep into rectifiers: Surpassing human-level performance on imagenet classification}, in: \bibinfo{booktitle}{Proceedings of the IEEE international conference on computer vision}, pp. \bibinfo{pages}{1026--1034}.
\bibitem[{He et~al.(2016)He, Zhang, Ren and Sun}]{he2016preactivResnet}
\bibinfo{author}{He, K.}, \bibinfo{author}{Zhang, X.}, \bibinfo{author}{Ren, S.}, \bibinfo{author}{Sun, J.}, \bibinfo{year}{2016}.
\newblock \bibinfo{title}{Identity mappings in deep residual networks}, in: \bibinfo{editor}{Leibe, B.}, \bibinfo{editor}{Matas, J.}, \bibinfo{editor}{Sebe, N.}, \bibinfo{editor}{Welling, M.} (Eds.), \bibinfo{booktitle}{Computer Vision -- ECCV 2016}, \bibinfo{publisher}{Springer International Publishing}, \bibinfo{address}{Cham}. pp. \bibinfo{pages}{630--645}.
\bibitem[{Heckemann et~al.(2003)Heckemann, Hartkens, Leung, Zheng, Hill, Hajnal and Rueckert}]{heckemann2003ixi}
\bibinfo{author}{Heckemann, R.A.}, \bibinfo{author}{Hartkens, T.}, \bibinfo{author}{Leung, K.K.}, \bibinfo{author}{Zheng, Y.}, \bibinfo{author}{Hill, D.L.}, \bibinfo{author}{Hajnal, J.V.}, \bibinfo{author}{Rueckert, D.}, \bibinfo{year}{2003}.
\newblock \bibinfo{title}{Information extraction from medical images: developing an e-science application based on the globus toolkit}, in: \bibinfo{booktitle}{Proceedings of the 2nd UK e-Science All Hands Meeting}.
\bibitem[{Heiliger et~al.(2023)Heiliger, Sekuboyina, Menze, Egger and Kleesiek}]{heiliger2023beyond}
\bibinfo{author}{Heiliger, L.}, \bibinfo{author}{Sekuboyina, A.}, \bibinfo{author}{Menze, B.}, \bibinfo{author}{Egger, J.}, \bibinfo{author}{Kleesiek, J.}, \bibinfo{year}{2023}.
\newblock \bibinfo{title}{Beyond medical imaging-a review of multimodal deep learning in radiology}.
\newblock \bibinfo{journal}{Authorea Preprints} .
\bibitem[{Huang et~al.(2020)Huang, Pareek, Seyyedi, Banerjee and Lungren}]{Huang2020survey}
\bibinfo{author}{Huang, S.C.}, \bibinfo{author}{Pareek, A.}, \bibinfo{author}{Seyyedi, S.}, \bibinfo{author}{Banerjee, I.}, \bibinfo{author}{Lungren, M.P.}, \bibinfo{year}{2020}.
\newblock \bibinfo{title}{Fusion of medical imaging and electronic health records using deep learning: a systematic review and implementation guidelines}.
\newblock \bibinfo{journal}{NPJ digital medicine} \bibinfo{volume}{3}, \bibinfo{pages}{136}.
\bibitem[{Iglesias et~al.(2011)Iglesias, Liu, Thompson and Tu}]{Iglesias2011Robex}
\bibinfo{author}{Iglesias, J.E.}, \bibinfo{author}{Liu, C.Y.}, \bibinfo{author}{Thompson, P.M.}, \bibinfo{author}{Tu, Z.}, \bibinfo{year}{2011}.
\newblock \bibinfo{title}{Robust brain extraction across datasets and comparison with publicly available methods}.
\newblock \bibinfo{journal}{IEEE Transactions on Medical Imaging} \bibinfo{volume}{30}, \bibinfo{pages}{1617--1634}.
\bibitem[{Jack~Jr et~al.(2008)Jack~Jr, Bernstein, Fox, Thompson, Alexander, Harvey, Borowski, Britson, L.~Whitwell, Ward et~al.}]{jack2008ADNI}
\bibinfo{author}{Jack~Jr, C.R.}, \bibinfo{author}{Bernstein, M.A.}, \bibinfo{author}{Fox, N.C.}, \bibinfo{author}{Thompson, P.}, \bibinfo{author}{Alexander, G.}, \bibinfo{author}{Harvey, D.}, \bibinfo{author}{Borowski, B.}, \bibinfo{author}{Britson, P.J.}, \bibinfo{author}{L.~Whitwell, J.}, \bibinfo{author}{Ward, C.}, et~al., \bibinfo{year}{2008}.
\newblock \bibinfo{title}{The alzheimer's disease neuroimaging initiative (adni): Mri methods}.
\newblock \bibinfo{journal}{Journal of Magnetic Resonance Imaging: An Official Journal of the International Society for Magnetic Resonance in Medicine} \bibinfo{volume}{27}, \bibinfo{pages}{685--691}.
\bibitem[{Jenkinson et~al.(2012)Jenkinson, Beckmann, Behrens, Woolrich and Smith}]{Jenkinson2012FSL}
\bibinfo{author}{Jenkinson, M.}, \bibinfo{author}{Beckmann, C.F.}, \bibinfo{author}{Behrens, T.E.}, \bibinfo{author}{Woolrich, M.W.}, \bibinfo{author}{Smith, S.M.}, \bibinfo{year}{2012}.
\newblock \bibinfo{title}{Fsl}.
\newblock \bibinfo{journal}{NeuroImage} \bibinfo{volume}{62}, \bibinfo{pages}{782--790}.
\newblock \bibinfo{note}{20 YEARS OF fMRI}.
\bibitem[{Lee et~al.(2022)Lee, Burkett, Min, Senjem, Lundt, Botha, Graff-Radford, Barnard, Gunter, Schwarz et~al.}]{lee2022BrainAgePrediction}
\bibinfo{author}{Lee, J.}, \bibinfo{author}{Burkett, B.J.}, \bibinfo{author}{Min, H.K.}, \bibinfo{author}{Senjem, M.L.}, \bibinfo{author}{Lundt, E.S.}, \bibinfo{author}{Botha, H.}, \bibinfo{author}{Graff-Radford, J.}, \bibinfo{author}{Barnard, L.R.}, \bibinfo{author}{Gunter, J.L.}, \bibinfo{author}{Schwarz, C.G.}, et~al., \bibinfo{year}{2022}.
\newblock \bibinfo{title}{Deep learning-based brain age prediction in normal aging and dementia}.
\newblock \bibinfo{journal}{Nature Aging} \bibinfo{volume}{2}, \bibinfo{pages}{412--424}.
\bibitem[{Letenneur et~al.(1999)Letenneur, Gilleron, Commenges, Helmer, Orgogozo and Dartigues}]{Letenneur1999SexAgeEducation_AD}
\bibinfo{author}{Letenneur, L.}, \bibinfo{author}{Gilleron, V.}, \bibinfo{author}{Commenges, D.}, \bibinfo{author}{Helmer, C.}, \bibinfo{author}{Orgogozo, J.M.}, \bibinfo{author}{Dartigues, J.F.}, \bibinfo{year}{1999}.
\newblock \bibinfo{title}{Are sex and educational level independent predictors of dementia and alzheimer{\textquoteright}s disease? incidence data from the paquid project}.
\newblock \bibinfo{journal}{Journal of Neurology, Neurosurgery \& Psychiatry} \bibinfo{volume}{66}, \bibinfo{pages}{177--183}.
\bibitem[{Levakov et~al.(2020)Levakov, Rosenthal, Shelef, Riklin~Raviv and Avidan}]{levakov2020brainage}
\bibinfo{author}{Levakov, G.}, \bibinfo{author}{Rosenthal, G.}, \bibinfo{author}{Shelef, I.}, \bibinfo{author}{Riklin~Raviv, T.}, \bibinfo{author}{Avidan, G.}, \bibinfo{year}{2020}.
\newblock \bibinfo{title}{From a deep learning model back to the brain—identifying regional predictors and their relation to aging}.
\newblock \bibinfo{journal}{Human Brain Mapping} \bibinfo{volume}{41}, \bibinfo{pages}{3235--3252}.
\bibitem[{Littwin and Wolf(2019)}]{littwin2019hyper3Dreconst}
\bibinfo{author}{Littwin, G.}, \bibinfo{author}{Wolf, L.}, \bibinfo{year}{2019}.
\newblock \bibinfo{title}{Deep meta functionals for shape representation}, in: \bibinfo{booktitle}{Proceedings of the IEEE/CVF International Conference on Computer Vision}, pp. \bibinfo{pages}{1824--1833}.
\bibitem[{Liu et~al.(2018)Liu, Zhang, Adeli and Shen}]{liu2018ADfusion_concat}
\bibinfo{author}{Liu, M.}, \bibinfo{author}{Zhang, J.}, \bibinfo{author}{Adeli, E.}, \bibinfo{author}{Shen, D.}, \bibinfo{year}{2018}.
\newblock \bibinfo{title}{Joint classification and regression via deep multi-task multi-channel learning for {A}lzheimer's disease diagnosis}.
\newblock \bibinfo{journal}{IEEE Transactions on Biomedical Engineering} \bibinfo{volume}{66}, \bibinfo{pages}{1195--1206}.
\bibitem[{Liu et~al.(2017)Liu, Wei, Chen, Yang, Meng, Wu, Bi, Zhang, Zuo and Qiu}]{liu2017SLIM}
\bibinfo{author}{Liu, W.}, \bibinfo{author}{Wei, D.}, \bibinfo{author}{Chen, Q.}, \bibinfo{author}{Yang, W.}, \bibinfo{author}{Meng, J.}, \bibinfo{author}{Wu, G.}, \bibinfo{author}{Bi, T.}, \bibinfo{author}{Zhang, Q.}, \bibinfo{author}{Zuo, X.N.}, \bibinfo{author}{Qiu, J.}, \bibinfo{year}{2017}.
\newblock \bibinfo{title}{Longitudinal test-retest neuroimaging data from healthy young adults in southwest china}.
\newblock \bibinfo{journal}{Scientific data} \bibinfo{volume}{4}, \bibinfo{pages}{1--9}.
\bibitem[{Lutati and Wolf(2021)}]{lutati2021hyperhyper}
\bibinfo{author}{Lutati, S.}, \bibinfo{author}{Wolf, L.}, \bibinfo{year}{2021}.
\newblock \bibinfo{title}{Hyperhypernetwork for the design of antenna arrays}, in: \bibinfo{booktitle}{International Conference on Machine Learning}, \bibinfo{organization}{PMLR}. pp. \bibinfo{pages}{7214--7223}.
\bibitem[{Marcus et~al.(2010)Marcus, Fotenos, Csernansky, Morris and Buckner}]{marcus2010OASIS}
\bibinfo{author}{Marcus, D.S.}, \bibinfo{author}{Fotenos, A.F.}, \bibinfo{author}{Csernansky, J.G.}, \bibinfo{author}{Morris, J.C.}, \bibinfo{author}{Buckner, R.L.}, \bibinfo{year}{2010}.
\newblock \bibinfo{title}{Open access series of imaging studies: longitudinal mri data in nondemented and demented older adults}.
\newblock \bibinfo{journal}{Journal of cognitive neuroscience} \bibinfo{volume}{22}, \bibinfo{pages}{2677--2684}.
\bibitem[{Marcus et~al.(2007)Marcus, Wang, Parker, Csernansky, Morris and Buckner}]{marcus2007OASIS}
\bibinfo{author}{Marcus, D.S.}, \bibinfo{author}{Wang, T.H.}, \bibinfo{author}{Parker, J.}, \bibinfo{author}{Csernansky, J.G.}, \bibinfo{author}{Morris, J.C.}, \bibinfo{author}{Buckner, R.L.}, \bibinfo{year}{2007}.
\newblock \bibinfo{title}{Open access series of imaging studies (oasis): cross-sectional mri data in young, middle aged, nondemented, and demented older adults}.
\newblock \bibinfo{journal}{Journal of cognitive neuroscience} \bibinfo{volume}{19}, \bibinfo{pages}{1498--1507}.
\bibitem[{Marek et~al.(2011)Marek, Jennings, Lasch, Siderowf, Tanner, Simuni, Coffey, Kieburtz, Flagg, Chowdhury, Poewe, Mollenhauer, Klinik, Sherer, Frasier, Meunier, Rudolph, Casaceli, Seibyl, Mendick, Schuff, Zhang, Toga, Crawford, Ansbach, {De Blasio}, Piovella, Trojanowski, Shaw, Singleton, Hawkins, Eberling, Brooks, Russell, Leary, Factor, Sommerfeld, Hogarth, Pighetti, Williams, Standaert, Guthrie, Hauser, Delgado, Jankovic, Hunter, Stern, Tran, Leverenz, Baca, Frank, Thomas, Richard, Deeley, Rees, Sprenger, Lang, Shill, Obradov, Fernandez, Winters, Berg, Gauss, Galasko, Fontaine, Mari, Gerstenhaber, Brooks, Malloy, Barone, Longo, Comery, Ravina, Grachev, Gallagher, Collins, Widnell, Ostrowizki, Fontoura, Ho, Luthman, van~der Brug, Reith and Taylor}]{MAREK2011PPMI}
\bibinfo{author}{Marek, K.}, \bibinfo{author}{Jennings, D.}, \bibinfo{author}{Lasch, S.}, \bibinfo{author}{Siderowf, A.}, \bibinfo{author}{Tanner, C.}, \bibinfo{author}{Simuni, T.}, \bibinfo{author}{Coffey, C.}, \bibinfo{author}{Kieburtz, K.}, \bibinfo{author}{Flagg, E.}, \bibinfo{author}{Chowdhury, S.}, \bibinfo{author}{Poewe, W.}, \bibinfo{author}{Mollenhauer, B.}, \bibinfo{author}{Klinik, P.E.}, \bibinfo{author}{Sherer, T.}, \bibinfo{author}{Frasier, M.}, \bibinfo{author}{Meunier, C.}, \bibinfo{author}{Rudolph, A.}, \bibinfo{author}{Casaceli, C.}, \bibinfo{author}{Seibyl, J.}, \bibinfo{author}{Mendick, S.}, \bibinfo{author}{Schuff, N.}, \bibinfo{author}{Zhang, Y.}, \bibinfo{author}{Toga, A.}, \bibinfo{author}{Crawford, K.}, \bibinfo{author}{Ansbach, A.}, \bibinfo{author}{{De Blasio}, P.}, \bibinfo{author}{Piovella, M.}, \bibinfo{author}{Trojanowski, J.}, \bibinfo{author}{Shaw, L.}, \bibinfo{author}{Singleton, A.}, \bibinfo{author}{Hawkins, K.}, \bibinfo{author}{Eberling, J.}, \bibinfo{author}{Brooks, D.},
  \bibinfo{author}{Russell, D.}, \bibinfo{author}{Leary, L.}, \bibinfo{author}{Factor, S.}, \bibinfo{author}{Sommerfeld, B.}, \bibinfo{author}{Hogarth, P.}, \bibinfo{author}{Pighetti, E.}, \bibinfo{author}{Williams, K.}, \bibinfo{author}{Standaert, D.}, \bibinfo{author}{Guthrie, S.}, \bibinfo{author}{Hauser, R.}, \bibinfo{author}{Delgado, H.}, \bibinfo{author}{Jankovic, J.}, \bibinfo{author}{Hunter, C.}, \bibinfo{author}{Stern, M.}, \bibinfo{author}{Tran, B.}, \bibinfo{author}{Leverenz, J.}, \bibinfo{author}{Baca, M.}, \bibinfo{author}{Frank, S.}, \bibinfo{author}{Thomas, C.A.}, \bibinfo{author}{Richard, I.}, \bibinfo{author}{Deeley, C.}, \bibinfo{author}{Rees, L.}, \bibinfo{author}{Sprenger, F.}, \bibinfo{author}{Lang, E.}, \bibinfo{author}{Shill, H.}, \bibinfo{author}{Obradov, S.}, \bibinfo{author}{Fernandez, H.}, \bibinfo{author}{Winters, A.}, \bibinfo{author}{Berg, D.}, \bibinfo{author}{Gauss, K.}, \bibinfo{author}{Galasko, D.}, \bibinfo{author}{Fontaine, D.}, \bibinfo{author}{Mari, Z.},
  \bibinfo{author}{Gerstenhaber, M.}, \bibinfo{author}{Brooks, D.}, \bibinfo{author}{Malloy, S.}, \bibinfo{author}{Barone, P.}, \bibinfo{author}{Longo, K.}, \bibinfo{author}{Comery, T.}, \bibinfo{author}{Ravina, B.}, \bibinfo{author}{Grachev, I.}, \bibinfo{author}{Gallagher, K.}, \bibinfo{author}{Collins, M.}, \bibinfo{author}{Widnell, K.L.}, \bibinfo{author}{Ostrowizki, S.}, \bibinfo{author}{Fontoura, P.}, \bibinfo{author}{Ho, T.}, \bibinfo{author}{Luthman, J.}, \bibinfo{author}{van~der Brug, M.}, \bibinfo{author}{Reith, A.D.}, \bibinfo{author}{Taylor, P.}, \bibinfo{year}{2011}.
\newblock \bibinfo{title}{The parkinson progression marker initiative (ppmi)}.
\newblock \bibinfo{journal}{Progress in Neurobiology} \bibinfo{volume}{95}, \bibinfo{pages}{629--635}.
\newblock \bibinfo{note}{Biological Markers for Neurodegenerative Diseases}.
\bibitem[{Mayer et~al.(2013)Mayer, Ruhl, Merideth, Ling, Hanlon, Bustillo and Canive}]{mayer2013COBRE}
\bibinfo{author}{Mayer, A.R.}, \bibinfo{author}{Ruhl, D.}, \bibinfo{author}{Merideth, F.}, \bibinfo{author}{Ling, J.}, \bibinfo{author}{Hanlon, F.M.}, \bibinfo{author}{Bustillo, J.}, \bibinfo{author}{Canive, J.}, \bibinfo{year}{2013}.
\newblock \bibinfo{title}{Functional imaging of the hemodynamic sensory gating response in schizophrenia}.
\newblock \bibinfo{journal}{Human brain mapping} \bibinfo{volume}{34}, \bibinfo{pages}{2302--2312}.
\bibitem[{Mazziotta et~al.(1995)Mazziotta, Toga, Evans, Fox and Lancaster}]{MAZZIOTTA1995ICBM}
\bibinfo{author}{Mazziotta, J.C.}, \bibinfo{author}{Toga, A.W.}, \bibinfo{author}{Evans, A.}, \bibinfo{author}{Fox, P.}, \bibinfo{author}{Lancaster, J.}, \bibinfo{year}{1995}.
\newblock \bibinfo{title}{A probabilistic atlas of the human brain: Theory and rationale for its development: The international consortium for brain mapping (icbm)}.
\newblock \bibinfo{journal}{NeuroImage} \bibinfo{volume}{2}, \bibinfo{pages}{89--101}.
\bibitem[{Menze et~al.(2014)Menze, Jakab, Bauer, Kalpathy-Cramer, Farahani, Kirby, Burren, Porz, Slotboom, Wiest et~al.}]{menze2014multimodal}
\bibinfo{author}{Menze, B.H.}, \bibinfo{author}{Jakab, A.}, \bibinfo{author}{Bauer, S.}, \bibinfo{author}{Kalpathy-Cramer, J.}, \bibinfo{author}{Farahani, K.}, \bibinfo{author}{Kirby, J.}, \bibinfo{author}{Burren, Y.}, \bibinfo{author}{Porz, N.}, \bibinfo{author}{Slotboom, J.}, \bibinfo{author}{Wiest, R.}, et~al., \bibinfo{year}{2014}.
\newblock \bibinfo{title}{The multimodal brain tumor image segmentation benchmark (brats)}.
\newblock \bibinfo{journal}{IEEE transactions on medical imaging} \bibinfo{volume}{34}, \bibinfo{pages}{1993--2024}.
\bibitem[{Nooner et~al.(2012)Nooner, Colcombe, Tobe, Mennes, Benedict, Moreno, Panek, Brown, Zavitz, Li et~al.}]{nooner2012nkiRS}
\bibinfo{author}{Nooner, K.B.}, \bibinfo{author}{Colcombe, S.J.}, \bibinfo{author}{Tobe, R.H.}, \bibinfo{author}{Mennes, M.}, \bibinfo{author}{Benedict, M.M.}, \bibinfo{author}{Moreno, A.L.}, \bibinfo{author}{Panek, L.J.}, \bibinfo{author}{Brown, S.}, \bibinfo{author}{Zavitz, S.T.}, \bibinfo{author}{Li, Q.}, et~al., \bibinfo{year}{2012}.
\newblock \bibinfo{title}{The nki-rockland sample: a model for accelerating the pace of discovery science in psychiatry}.
\newblock \bibinfo{journal}{Frontiers in neuroscience} \bibinfo{volume}{6}, \bibinfo{pages}{152}.
\bibitem[{Pandeya and Lee(2021)}]{pandeya2021late_fusion_softmax}
\bibinfo{author}{Pandeya, Y.R.}, \bibinfo{author}{Lee, J.}, \bibinfo{year}{2021}.
\newblock \bibinfo{title}{Deep learning-based late fusion of multimodal information for emotion classification of music video}.
\newblock \bibinfo{journal}{Multimedia Tools and Applications} \bibinfo{volume}{80}, \bibinfo{pages}{2887--2905}.
\bibitem[{Peng et~al.(2021)Peng, Gong, Beckmann, Vedaldi and Smith}]{Peng2021lightDNN4BrainAge}
\bibinfo{author}{Peng, H.}, \bibinfo{author}{Gong, W.}, \bibinfo{author}{Beckmann, C.F.}, \bibinfo{author}{Vedaldi, A.}, \bibinfo{author}{Smith, S.M.}, \bibinfo{year}{2021}.
\newblock \bibinfo{title}{Accurate brain age prediction with lightweight deep neural networks}.
\newblock \bibinfo{journal}{Medical Image Analysis} \bibinfo{volume}{68}, \bibinfo{pages}{101871}.
\bibitem[{Perez et~al.(2018)Perez, Strub, de~Vries, Dumoulin and Courville}]{perez2018film}
\bibinfo{author}{Perez, E.}, \bibinfo{author}{Strub, F.}, \bibinfo{author}{de~Vries, H.}, \bibinfo{author}{Dumoulin, V.}, \bibinfo{author}{Courville, A.}, \bibinfo{year}{2018}.
\newblock \bibinfo{title}{{F}i{LM}: visual reasoning with a general conditioning layer}.
\newblock \bibinfo{journal}{Proceedings of the AAAI Conference on Artificial Intelligence} \bibinfo{volume}{32}.
\bibitem[{Pi{\c{c}}arra and Glocker(2023)}]{piccarra2023}
\bibinfo{author}{Pi{\c{c}}arra, C.}, \bibinfo{author}{Glocker, B.}, \bibinfo{year}{2023}.
\newblock \bibinfo{title}{Analysing race and sex bias in brain age prediction}, in: \bibinfo{booktitle}{Workshop on Clinical Image-Based Procedures}, \bibinfo{organization}{Springer}. pp. \bibinfo{pages}{194--204}.
\bibitem[{Pinel et~al.(2012)Pinel, Fauchereau, Moreno, Barbot, Lathrop, Zelenika, Le~Bihan, Poline, Bourgeron and Dehaene}]{pinel201257Brainomics}
\bibinfo{author}{Pinel, P.}, \bibinfo{author}{Fauchereau, F.}, \bibinfo{author}{Moreno, A.}, \bibinfo{author}{Barbot, A.}, \bibinfo{author}{Lathrop, M.}, \bibinfo{author}{Zelenika, D.}, \bibinfo{author}{Le~Bihan, D.}, \bibinfo{author}{Poline, J.B.}, \bibinfo{author}{Bourgeron, T.}, \bibinfo{author}{Dehaene, S.}, \bibinfo{year}{2012}.
\newblock \bibinfo{title}{Genetic variants of foxp2 and kiaa0319/ttrap/them2 locus are associated with altered brain activation in distinct language-related regions}.
\newblock \bibinfo{journal}{Journal of Neuroscience} \bibinfo{volume}{32}, \bibinfo{pages}{817--825}.
\bibitem[{Poldrack et~al.(2016)Poldrack, Congdon, Triplett, Gorgolewski, Karlsgodt, Mumford, Sabb, Freimer, London, Cannon et~al.}]{poldrack2016CNP}
\bibinfo{author}{Poldrack, R.A.}, \bibinfo{author}{Congdon, E.}, \bibinfo{author}{Triplett, W.}, \bibinfo{author}{Gorgolewski, K.}, \bibinfo{author}{Karlsgodt, K.}, \bibinfo{author}{Mumford, J.}, \bibinfo{author}{Sabb, F.}, \bibinfo{author}{Freimer, N.}, \bibinfo{author}{London, E.}, \bibinfo{author}{Cannon, T.}, et~al., \bibinfo{year}{2016}.
\newblock \bibinfo{title}{A phenome-wide examination of neural and cognitive function}.
\newblock \bibinfo{journal}{Scientific data} \bibinfo{volume}{3}, \bibinfo{pages}{1--12}.
\bibitem[{Prabhu et~al.(2022)Prabhu, Berkebile, Rajagopalan, Yao, Shi, Giuste, Zhong, Sun and Wang}]{prabhu2022multimodal_AD_entropydecision}
\bibinfo{author}{Prabhu, S.S.}, \bibinfo{author}{Berkebile, J.A.}, \bibinfo{author}{Rajagopalan, N.}, \bibinfo{author}{Yao, R.}, \bibinfo{author}{Shi, W.}, \bibinfo{author}{Giuste, F.}, \bibinfo{author}{Zhong, Y.}, \bibinfo{author}{Sun, J.}, \bibinfo{author}{Wang, M.D.}, \bibinfo{year}{2022}.
\newblock \bibinfo{title}{Multi-modal deep learning models for {A}lzheimer's disease prediction using {MRI} and {EHR}}, in: \bibinfo{booktitle}{2022 IEEE 22nd International Conference on Bioinformatics and Bioengineering (BIBE)}, \bibinfo{organization}{IEEE}. pp. \bibinfo{pages}{168--173}.
\bibitem[{Qiu et~al.(2018)Qiu, Chang, Panagia, Gopal, Au and Kolachalama}]{Qiu2018ADfusion_cognitiveTests}
\bibinfo{author}{Qiu, S.}, \bibinfo{author}{Chang, G.H.}, \bibinfo{author}{Panagia, M.}, \bibinfo{author}{Gopal, D.M.}, \bibinfo{author}{Au, R.}, \bibinfo{author}{Kolachalama, V.B.}, \bibinfo{year}{2018}.
\newblock \bibinfo{title}{Fusion of deep learning models of {MRI} scans, mini–mental state examination, and logical memory test enhances diagnosis of mild cognitive impairment}.
\newblock \bibinfo{journal}{Alzheimer's and Dementia: Diagnosis, Assessment and Disease Monitoring} \bibinfo{volume}{10}, \bibinfo{pages}{737--749}.
\bibitem[{Radford et~al.(2021)Radford, Kim, Hallacy, Ramesh, Goh, Agarwal, Sastry, Askell, Mishkin, Clark et~al.}]{radford2021CLIP}
\bibinfo{author}{Radford, A.}, \bibinfo{author}{Kim, J.W.}, \bibinfo{author}{Hallacy, C.}, \bibinfo{author}{Ramesh, A.}, \bibinfo{author}{Goh, G.}, \bibinfo{author}{Agarwal, S.}, \bibinfo{author}{Sastry, G.}, \bibinfo{author}{Askell, A.}, \bibinfo{author}{Mishkin, P.}, \bibinfo{author}{Clark, J.}, et~al., \bibinfo{year}{2021}.
\newblock \bibinfo{title}{Learning transferable visual models from natural language supervision}, in: \bibinfo{booktitle}{International conference on machine learning}, \bibinfo{organization}{PMLR}. pp. \bibinfo{pages}{8748--8763}.
\bibitem[{Rao et~al.(2022)Rao, Ganaraja, Murlimanju, Joy, Krishnamurthy and Agrawal}]{Rao2022}
\bibinfo{author}{Rao, Y.}, \bibinfo{author}{Ganaraja, B.}, \bibinfo{author}{Murlimanju, B.}, \bibinfo{author}{Joy, T.}, \bibinfo{author}{Krishnamurthy, A.}, \bibinfo{author}{Agrawal, A.}, \bibinfo{year}{2022}.
\newblock \bibinfo{title}{Hippocampus and its involvement in {A}lzheimer's disease: a review}.
\newblock \bibinfo{journal}{3 Biotech} \bibinfo{volume}{12}.
\bibitem[{Reinhold et~al.(2018)Reinhold, Dewey, Carass and Prince}]{Reinhold2018}
\bibinfo{author}{Reinhold, J.C.}, \bibinfo{author}{Dewey, B.E.}, \bibinfo{author}{Carass, A.}, \bibinfo{author}{Prince, J.L.}, \bibinfo{year}{2018}.
\newblock \bibinfo{title}{Evaluating the impact of intensity normalization on mr image synthesis}.
\newblock \bibinfo{journal}{CoRR} \bibinfo{volume}{abs/1812.04652}.
\newblock \href{http://arxiv.org/abs/1812.04652}{\tt arXiv:1812.04652}.
\bibitem[{Salta et~al.(2023)Salta, Lazarov, Fitzsimons, Tanzi, Lucassen and Choi}]{Salta2023}
\bibinfo{author}{Salta, E.}, \bibinfo{author}{Lazarov, O.}, \bibinfo{author}{Fitzsimons, P.}, \bibinfo{author}{Tanzi, R.}, \bibinfo{author}{Lucassen, P.}, \bibinfo{author}{Choi, S.}, \bibinfo{year}{2023}.
\newblock \bibinfo{title}{Adult hippocampal neurogenesis in {A}lzheimer’s disease: A roadmap to clinical relevance}.
\newblock \bibinfo{journal}{Cell Stem Cell} \bibinfo{volume}{30}, \bibinfo{pages}{120--136}.
\bibitem[{Shafto et~al.(2014)Shafto, Tyler, Dixon, Taylor, Rowe, Cusack, Calder, Marslen-Wilson, Duncan, Dalgleish et~al.}]{shafto2014CamCAN}
\bibinfo{author}{Shafto, M.A.}, \bibinfo{author}{Tyler, L.K.}, \bibinfo{author}{Dixon, M.}, \bibinfo{author}{Taylor, J.R.}, \bibinfo{author}{Rowe, J.B.}, \bibinfo{author}{Cusack, R.}, \bibinfo{author}{Calder, A.J.}, \bibinfo{author}{Marslen-Wilson, W.D.}, \bibinfo{author}{Duncan, J.}, \bibinfo{author}{Dalgleish, T.}, et~al., \bibinfo{year}{2014}.
\newblock \bibinfo{title}{The cambridge centre for ageing and neuroscience (cam-can) study protocol: a cross-sectional, lifespan, multidisciplinary examination of healthy cognitive ageing}.
\newblock \bibinfo{journal}{BMC neurology} \bibinfo{volume}{14}, \bibinfo{pages}{1--25}.
\bibitem[{Simonyan and Zisserman(2014)}]{simonyan2014VGG}
\bibinfo{author}{Simonyan, K.}, \bibinfo{author}{Zisserman, A.}, \bibinfo{year}{2014}.
\newblock \bibinfo{title}{Very deep convolutional networks for large-scale image recognition}.
\newblock \bibinfo{journal}{arXiv preprint arXiv:1409.1556} .
\bibitem[{Spasov et~al.(2018)Spasov, Passamonti, Duggento, Liò and Toschi}]{Simeon2018_2classADfusion}
\bibinfo{author}{Spasov, S.E.}, \bibinfo{author}{Passamonti, L.}, \bibinfo{author}{Duggento, A.}, \bibinfo{author}{Liò, P.}, \bibinfo{author}{Toschi, N.}, \bibinfo{year}{2018}.
\newblock \bibinfo{title}{A multi-modal convolutional neural network framework for the prediction of {A}lzheimer’s disease}, in: \bibinfo{booktitle}{2018 40th Annual International Conference of the IEEE Engineering in Medicine and Biology Society (EMBC)}, pp. \bibinfo{pages}{1271--1274}.
\bibitem[{Sudlow et~al.(2015)Sudlow, Gallacher, Allen, Beral, Burton, Danesh, Downey, Elliott, Green, Landray et~al.}]{sudlow2015UK_Biobank}
\bibinfo{author}{Sudlow, C.}, \bibinfo{author}{Gallacher, J.}, \bibinfo{author}{Allen, N.}, \bibinfo{author}{Beral, V.}, \bibinfo{author}{Burton, P.}, \bibinfo{author}{Danesh, J.}, \bibinfo{author}{Downey, P.}, \bibinfo{author}{Elliott, P.}, \bibinfo{author}{Green, J.}, \bibinfo{author}{Landray, M.}, et~al., \bibinfo{year}{2015}.
\newblock \bibinfo{title}{Uk biobank: an open access resource for identifying the causes of a wide range of complex diseases of middle and old age}.
\newblock \bibinfo{journal}{PLoS medicine} \bibinfo{volume}{12}, \bibinfo{pages}{e1001779}.
\bibitem[{Sui et~al.(2023)Sui, Zhi and Calhoun}]{sui2023data}
\bibinfo{author}{Sui, J.}, \bibinfo{author}{Zhi, D.}, \bibinfo{author}{Calhoun, V.D.}, \bibinfo{year}{2023}.
\newblock \bibinfo{title}{Data-driven multimodal fusion: approaches and applications in psychiatric research}.
\newblock \bibinfo{journal}{Psychoradiology} , \bibinfo{pages}{kkad026}.
\bibitem[{Tustison et~al.(2010)Tustison, Avants, Cook, Zheng, Egan, Yushkevich and Gee}]{Tustison2010N4ITK}
\bibinfo{author}{Tustison, N.J.}, \bibinfo{author}{Avants, B.B.}, \bibinfo{author}{Cook, P.A.}, \bibinfo{author}{Zheng, Y.}, \bibinfo{author}{Egan, A.}, \bibinfo{author}{Yushkevich, P.A.}, \bibinfo{author}{Gee, J.C.}, \bibinfo{year}{2010}.
\newblock \bibinfo{title}{N4itk: Improved n3 bias correction}.
\newblock \bibinfo{journal}{IEEE Transactions on Medical Imaging} \bibinfo{volume}{29}, \bibinfo{pages}{1310--1320}.
\bibitem[{Venugopalan et~al.(2021)Venugopalan, Tong, Hassanzadeh and Wang}]{venugopalan2021ADfusion_img_tab_snp}
\bibinfo{author}{Venugopalan, J.}, \bibinfo{author}{Tong, L.}, \bibinfo{author}{Hassanzadeh, H.R.}, \bibinfo{author}{Wang, M.D.}, \bibinfo{year}{2021}.
\newblock \bibinfo{title}{Multimodal deep learning models for early detection of {A}lzheimer’s disease stage}.
\newblock \bibinfo{journal}{Scientific reports} \bibinfo{volume}{11}, \bibinfo{pages}{3254}.
\bibitem[{Wang et~al.(2023)Wang, Liu, Ko, Wells, Berkowitz, Horng and Golland}]{wang2023sample}
\bibinfo{author}{Wang, P.}, \bibinfo{author}{Liu, Y.}, \bibinfo{author}{Ko, C.Y.}, \bibinfo{author}{Wells, W.M.}, \bibinfo{author}{Berkowitz, S.}, \bibinfo{author}{Horng, S.}, \bibinfo{author}{Golland, P.}, \bibinfo{year}{2023}.
\newblock \bibinfo{title}{Sample-specific debiasing for better image-text models}, in: \bibinfo{booktitle}{Machine Learning for Healthcare Conference}, pp. \bibinfo{pages}{788--803}.
\bibitem[{Wang et~al.(2018)Wang, Phillips, Sui, Liu, Yang and Cheng}]{wang2018CN_ADclass}
\bibinfo{author}{Wang, S.H.}, \bibinfo{author}{Phillips, P.}, \bibinfo{author}{Sui, Y.}, \bibinfo{author}{Liu, B.}, \bibinfo{author}{Yang, M.}, \bibinfo{author}{Cheng, H.}, \bibinfo{year}{2018}.
\newblock \bibinfo{title}{Classification of {A}lzheimer’s disease based on eight-layer convolutional neural network with leaky rectified linear unit and max pooling}.
\newblock \bibinfo{journal}{Journal of medical systems} \bibinfo{volume}{42}, \bibinfo{pages}{1--11}.
\bibitem[{Wen et~al.(2020)Wen, Thibeau-Sutre, Diaz-Melo, Samper-Gonz{\'a}lez, Routier, Bottani, Dormont, Durrleman, Burgos, Colliot et~al.}]{wen2020CNN_AD_overview}
\bibinfo{author}{Wen, J.}, \bibinfo{author}{Thibeau-Sutre, E.}, \bibinfo{author}{Diaz-Melo, M.}, \bibinfo{author}{Samper-Gonz{\'a}lez, J.}, \bibinfo{author}{Routier, A.}, \bibinfo{author}{Bottani, S.}, \bibinfo{author}{Dormont, D.}, \bibinfo{author}{Durrleman, S.}, \bibinfo{author}{Burgos, N.}, \bibinfo{author}{Colliot, O.}, et~al., \bibinfo{year}{2020}.
\newblock \bibinfo{title}{Convolutional neural networks for classification of {A}lzheimer's disease: Overview and reproducible evaluation}.
\newblock \bibinfo{journal}{Medical image analysis} \bibinfo{volume}{63}, \bibinfo{pages}{101694}.
\bibitem[{Wolf et~al.(2022)Wolf, Pölsterl and Wachinger}]{wolf2022DAFT}
\bibinfo{author}{Wolf, T.N.}, \bibinfo{author}{Pölsterl, S.}, \bibinfo{author}{Wachinger, C.}, \bibinfo{year}{2022}.
\newblock \bibinfo{title}{{DAFT}: A universal module to interweave tabular data and {3D} images in {CNN}s}.
\newblock \bibinfo{journal}{NeuroImage} \bibinfo{volume}{260}, \bibinfo{pages}{119505}.
\bibitem[{Wydma{\'n}ski et~al.(2023)Wydma{\'n}ski, Bulenok and {\'S}mieja}]{wydmanski2023hypertab}
\bibinfo{author}{Wydma{\'n}ski, W.}, \bibinfo{author}{Bulenok, O.}, \bibinfo{author}{{\'S}mieja, M.}, \bibinfo{year}{2023}.
\newblock \bibinfo{title}{Hyper{T}ab: Hypernetwork approach for deep learning on small tabular datasets}.
\newblock \bibinfo{journal}{arXiv preprint arXiv:2304.03543} .
\bibitem[{Zhou et~al.(2019)Zhou, Thung, Zhu and Shen}]{zhou2019ADfusion_MRI_PET_genetic}
\bibinfo{author}{Zhou, T.}, \bibinfo{author}{Thung, K.H.}, \bibinfo{author}{Zhu, X.}, \bibinfo{author}{Shen, D.}, \bibinfo{year}{2019}.
\newblock \bibinfo{title}{Effective feature learning and fusion of multimodality data using stage-wise deep neural network for dementia diagnosis}.
\newblock \bibinfo{journal}{Human brain mapping} \bibinfo{volume}{40}, \bibinfo{pages}{1001--1016}.
\bibitem[{Zuo et~al.(2014)Zuo, Anderson, Bellec, Birn, Biswal, Blautzik, Breitner, Buckner, Calhoun, Castellanos et~al.}]{zuo2014CoRR}
\bibinfo{author}{Zuo, X.N.}, \bibinfo{author}{Anderson, J.S.}, \bibinfo{author}{Bellec, P.}, \bibinfo{author}{Birn, R.M.}, \bibinfo{author}{Biswal, B.B.}, \bibinfo{author}{Blautzik, J.}, \bibinfo{author}{Breitner, J.}, \bibinfo{author}{Buckner, R.L.}, \bibinfo{author}{Calhoun, V.D.}, \bibinfo{author}{Castellanos, F.X.}, et~al., \bibinfo{year}{2014}.
\newblock \bibinfo{title}{An open science resource for establishing reliability and reproducibility in functional connectomics}.
\newblock \bibinfo{journal}{Scientific data} \bibinfo{volume}{1}, \bibinfo{pages}{1--13}.

\end{thebibliography}


\appendix
\bigskip
\noindent
{\huge\textbf{Appendix}}

\section{Brain Age Prediction}  \label{sec:Brain-Age-Prediction}
\subsection{Datasets and Preprocessing} \label{sec:Brain-Age-dataset}
Table~\ref{tab:brainage_data} presents the datasets used for the brain age prediction task.
The preprocessing pipeline of the T1-weighted MRI scans is identical to the one proposed in~\citep{levakov2020brainage}, using Nipype~\citep{gorgolewski2011nipype}.
Initially, RobustFov~\cite{Jenkinson2012FSL} was employed to remove the neck and shoulders from each scan. Subsequently, Robex~\cite{Iglesias2011Robex} was utilized for brain extraction, followed by expanding the brain mask to encompass the CSF surrounding the brain. The correction for intensity non-uniformity was carried out using N4BiasFieldCorrection~\cite{Tustison2010N4ITK}. Further, intensity normalization was implemented using fuzzy c-means and WM-based mean normalization~\cite{Reinhold2018}. Ultimately, the images were resampled to a resolution of 1.75 mm$^3$ and cropped to a 90x120x99 voxels box centered around the brain mask's center of mass.

\begin{table*}[!h]
\footnotesize
\caption{\label{tab:brainage_data}
List of all datasets used for the brain age prediction task. For each dataset, the number of available subjects (N), the sex distribution, and the mean and std of the age distribution are provided.}
\centering
\begin{tabular}{|l|c|c|c|}
\hline

\textbf{Study/database} & \textbf{N} & \textbf{Age - mean (±std)} & \textbf{Sex (M:F)} \\ \hline
UK Biobank  \citep{sudlow2015UK_Biobank}                                      & 17,088 & 63.1 (±7.5) & 8,118:8,970 \\ \hline
Consortium for Reliability and Reproducibility, CoRR \citep{zuo2014CoRR}   & 1,350 & 26.1 (±15.9) & 661:689 \\ \hline
Alzheimer's Disease Neuroimaging Initiative, ADNI \citep{jack2008ADNI}     & 659 & 72.9 (±6.0) & 432:227 \\ \hline
Brain Genomics Superstruct Project, GSP \citep{DVN25833_2014GSP}              & 1,099 & 21.5 (±2.9) & 469:630 \\ \hline
Functional Connectomes Project, FCP \citep{biswal2010FCP}               & 922 & 28.2 (±13.9) & 393:529 \\ \hline
Autism Brain Imaging Data Exchange, ABIDE \citep{di2014ABIDE}            & 506 & 16.9 (±7.7) & 411:95 \\ \hline
Parkinson's Progression Markers Initiative, PPMI \citep{MAREK2011PPMI}   & 172 & 60.3 (±11.0) & 105:67 \\ \hline
International Consortium for Brain Mapping, ICBM \citep{MAZZIOTTA1995ICBM}  & 592 & 30.9 (±12.5) & 319:273 \\ \hline
Australian Imaging, Biomarkers and Lifestyle, AIBL \citep{ellis2009AIBL} & 604 & 73.0 (±6.6) & 268:335 \\ \hline
Southwest University LongitudinalImaging Multimodal, SLIM \citep{liu2017SLIM}     & 571 & 20.1 (±1.3) & 251:320 \\ \hline
Information extraction from Images, IXI \citep{heckemann2003ixi}           & 562 & 48.2 (±16.5) & 251:311 \\ \hline
Open Access Series of Imaging Studies, OASIS \citep{marcus2007OASIS, marcus2010OASIS}   & 401 & 51.7 (±24.9) & 144:257 \\ \hline
Consortium for Neuropsychiatric Phenomics, CNP \citep{poldrack2016CNP}    & 124 & 31.5 (±8.8) & 65:59 \\ \hline
Center for Biomedical Research Excellence, COBRE \citep{mayer2013COBRE}    & 74 & 35.8 (±11.6) & 51:23 \\ \hline
Child and Adolescent NeuroDevelopment Initiative, CANDI \citep{frazier2008CANDI}    & 29 & 10.0 (±2.9) & 17:12 \\ \hline
Brainomics \citep{pinel201257Brainomics}                                   & 68 & 25.4 (±7.1) & 31:37 \\ \hline
Cambridge Centre for Ageing and Neuroscience, CamCAN \citep{shafto2014CamCAN}    & 652 & 54.3 (±18.6) & 322:330 \\ \hline
Child Mind Institute - Healthy Brain Network, HBN \citep{alexander2017HBN}         & 304 & 10.0 (±3.7) & 164:140 \\ \hline
Nathan Kline Institute-Rockland Sample, NKI \citep{nooner2012nkiRS}       & 914 & 37.1 (±21.7) & 359:555 \\ \hline
Overall                                                        & 26,691 & 53.7 (±19.5) & 12,831:13,859 \\ \hline
\end{tabular}
\end{table*}

\textcolor{black}{\subsection{Brain Age Prediction Model Architecture} \label{sec:Brainage_parameters}
Table~\ref{tbl:BAModelArch} presents the architectural details of the proposed HyperFusion model for brain age prediction presented in Section~\ref{sec:method_brainage_using_hyper} of the main paper.}
\begin{table}
\centering
\textcolor{black}{
\footnotesize   
\caption{Architecture of the  brain age prediction model\label{tbl:BAModelArch}}
\begin{tabular}{|c|l|l|r|} 
\hline
\multicolumn{4}{|c|}{Brain Age Prediction Model Architecture} \\ \hline
\textbf{Index} & \textbf{Name} & \textbf{Type} & \textbf{Params} \\ \hline
0  & conv1\_a     & Conv3d      & 448    \\ \hline
1  & conv1\_b     & Conv3d      & 6.9 K  \\ \hline
2  & batchnorm1  & BatchNorm3d & 32     \\ \hline
3  & conv2\_a     & Conv3d      & 13.9 K \\ \hline
4  & conv2\_b     & Conv3d      & 27.7 K \\ \hline
5  & batchnorm2  & BatchNorm3d & 64     \\ \hline
6  & conv3\_a     & Conv3d      & 55.4 K \\ \hline
7  & conv3\_b     & Conv3d      & 110 K  \\ \hline
8  & batchnorm3  & BatchNorm3d & 128    \\ \hline
9  & dropout1    & Dropout3d   & 0      \\ \hline
10 & linear1     & HyperLinear & 1.3 M  \\ \hline
11 & linear2     & HyperLinear & 1.1 K  \\ \hline
12 & linear3     & HyperLinear & 4.2 K  \\ \hline
13 & final\_layer & HyperLinear & 133    \\ \hline
\multicolumn{4}{|c|}{} \\ \hline
\multicolumn{3}{|l|}{Trainable params} & 1.5 M \\ \hline
\multicolumn{3}{|l|}{Total params} & 1.5 M \\ \hline
\multicolumn{3}{|l|}{Total estimated model params size (MB)} & 5.929 \\ \hline
\end{tabular}}
\end{table}

\begin{figure*}[h]
    \centering
    \includegraphics[scale=0.85]{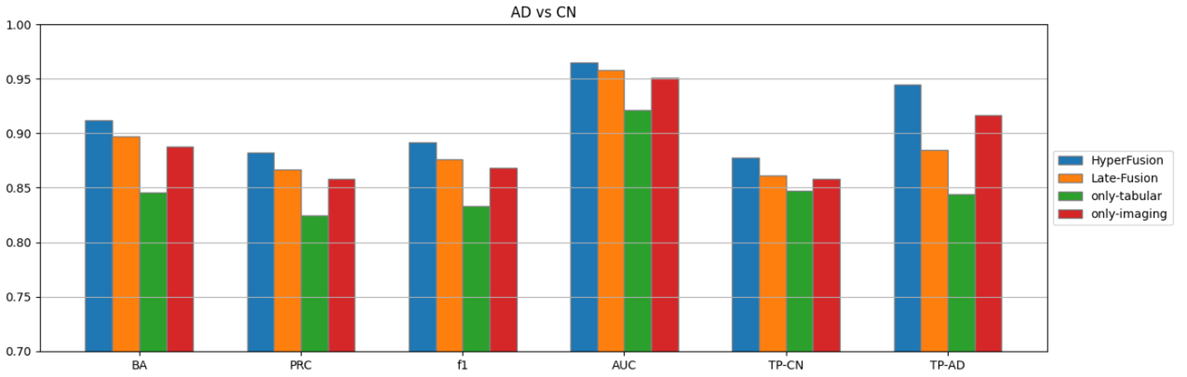}
    \caption{Binary classification experiments comparing imaging/tabular data single modality and Hyperfusion. 
    }
    \label{fig:_AD_vs_CN}
\end{figure*}

\textcolor{black}{\section{Binary AD classification}  \label{sec:AD_binary}} 
\textcolor{black}{
We adapted our hypernetwork framework which performs multi-class Alzheimer Disease (AD) classification to address binary classification, thus further demonstrating the strength of our method.
The binary classification task is considered less challenging since structural brain differences between AD patients and cognitively normal (CN) subjects are apparent. Nevertheless, there is still a significant gain in integrating both imaging and tabular data using the proposed hypernetwork framework on our sorted ADNI dataset, as is demonstrated in Figure~\ref{fig:_AD_vs_CN}. To further demonstrate the advantage of the proposed hypernetwork framework we fused the soft classification results of both the tabular-only and the imaging-only and present the resulting late-fusion predictions (orange bars) in the figure. 
While the late-fusion results are better than those obtained by the tabular-only for all metrics and imaging-only for most metrics they are inferior with respect to the hypernetwork performances.
}

\section{Multiclass AD classification}

\subsection{Data splitting, model training and evaluation}  \label{sec:AD_data_splitting}
\begin{figure}
    \centering
    \includegraphics[scale=0.45]{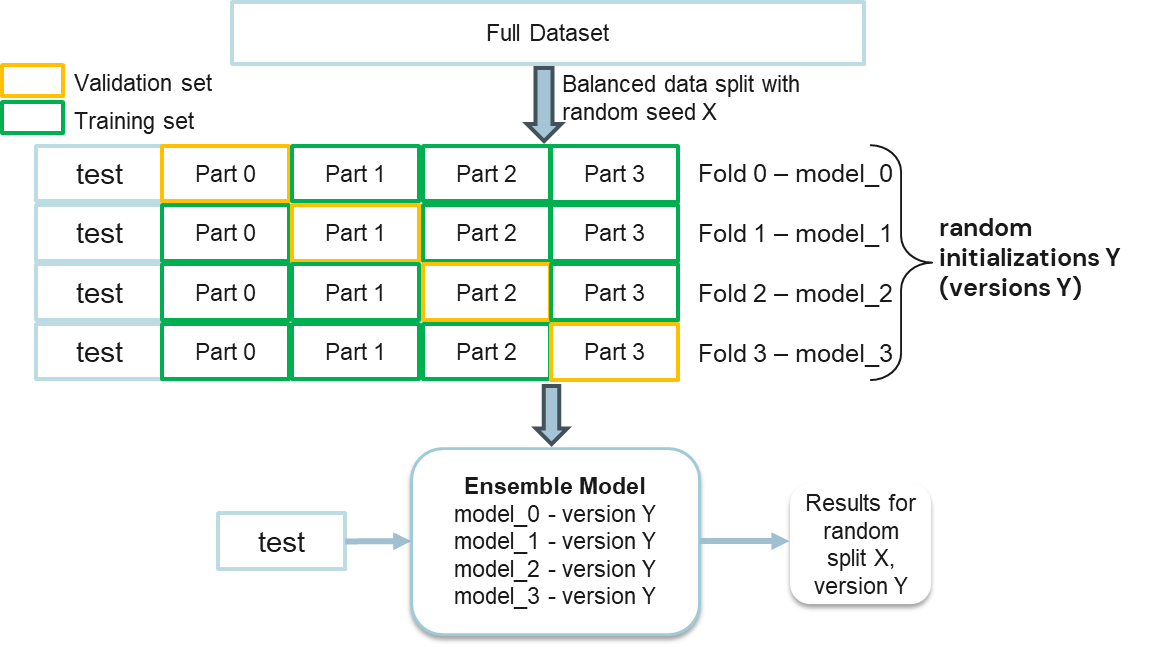}
    \caption{A schematic illustration of the data splitting and model training and evaluation for multiclass AD classification: each round is based on a different random initialization of a model. The entire process is performed using three random seeds with three versions for each, i.e., nine cross-validation experiments all together.}
    \label{fig:data_split}
\end{figure}

Figure~\ref{fig:data_split} depicts the partition of the ADNI data as well as the evaluation method for the cross validation experiments of the AD classification.

\textcolor{black}{\subsection{AD Classification Model Architecture}  \label{sec:AD_parameters}
Table~\ref{tbl:ADModelArch} presents the architectural details of the proposed HyperFusion model for AD classification presented in Section~\ref{sec:method_AD_using_hyper} of the main paper. }

\begin{table}
\textcolor{black}{
\caption{Architecture of the AD classification model \label{tbl:ADModelArch}}
\footnotesize   
\begin{tabular}{|c|l|l|r|}
\hline
\multicolumn{4}{|c|}{AD Classification Model Parameters} \\ \hline
\textbf{Index} & \textbf{Name} & \textbf{Type} & \textbf{Params} \\ \hline
0  & conv\_bn\_relu        & Sequential            & 480    \\ \hline
1  & max\_pool3d\_1        & MaxPool3d             & 0      \\ \hline
2  & block1              & PreactivResBlock\_bn   & 42.2 K \\ \hline
3  & block2              & PreactivResBlock\_bn   & 168 K  \\ \hline
4  & block3              & PreactivResBlock\_bn   & 672 K  \\ \hline
5  & block4              & HyperPreactivResBlock  & 3.0 M  \\ \hline
6  & adaptive\_avg\_pool3d & AdaptiveAvgPool3d     & 0      \\ \hline
7  & linear\_drop1        & Dropout               & 0      \\ \hline
8  & fc1                 & LinearLayer           & 16.4 K \\ \hline
9  & linear\_drop2        & Dropout               & 0      \\ \hline
10 & fc2                 & LinearLayer           & 195    \\ \hline
11 & relu                & ReLU                  & 0      \\ \hline
\multicolumn{4}{|c|}{} \\ \hline
\multicolumn{3}{|l|}{Trainable params} & 3.9 M \\ \hline
\multicolumn{3}{|l|}{Total params} & 3.9 M \\ \hline
\multicolumn{3}{|l|}{Total estimated model params size (MB)} & 15.415 \\ \hline
\end{tabular}}
\end{table}

\subsection{AD Classification Confusion Matrices}  \label{sec:AD_conf_matrix}
\begin{figure*}
    \centering
    \includegraphics[scale=0.75]{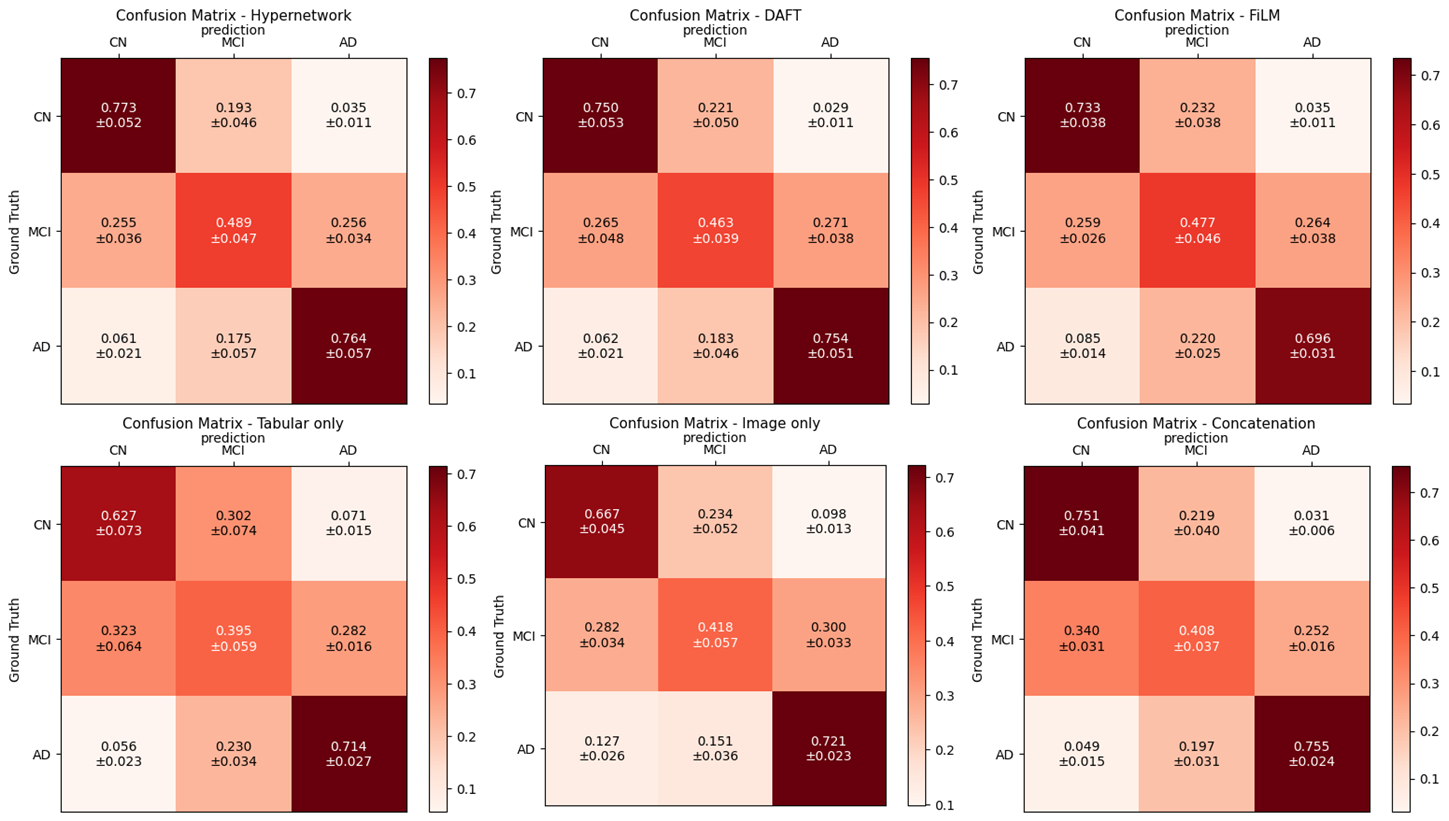}
    \caption{Confusion matrices for the AD classification task. Each matrix represents the predicted results versus the ground truth, with row-normalization for easier interpretation. Each cell in a matrix contains the mean and standard deviation values calculated over all conducted experiments.}
    \label{fig:conf_matrices}
\end{figure*}

Figure~\ref{fig:conf_matrices} presents the confusion matrices obtained by the proposed the hyperfusion model for the AD in comparison to other methods. The confusion matrices provide statistics on the the accurate and inaccurate classification rates of each class.

\subsection{Class Weights for the AD Classification Task}  \label{sec:AD_diff_class_weights}
\textcolor{black}{As mentioned in the main manuscript, we trained the hyperfusion framewotk for multi-class AD classification with weighted categorical cross entropy (WCE) loss, where the weights of the classes (CN, MCI, AD) are approximately inverse-proportional to their frequency. 
To assess our choice, we performed an ablation study comparing the proposed approach with a prevalent alternative to address data imbalance, an oversampling of the less frequent classes using standard CE loss. The results are presented in Table~\ref{tab:AD_WCEablation} showing the advantage of using the WCE loss.}
\begin{table*}[!h]
\footnotesize
\caption{\label{tab:AD_WCEablation}
\textcolor{black}{Hyperfusion performances using WCE/CE loss with/without upsampling}}
\centering
\textcolor{black}{
\begin{tabular}{|l|c|c|c|c|c|c|c|}
\hline
\textbf{model} & \textbf{BA} & \textbf{PRC} & \textbf{f1 macro} & \textbf{AUC macro} & \textbf{TP-CN} & \textbf{TP-MCI} & \textbf{TP-AD} \\ \hline
WCE loss  & \textbf{.673} & \textbf{.624} & \textbf{.630} & \textbf{.822} & \textbf{.759} & .495 & \textbf{.764}  \\ \hline
CE w upsample  & .643 & .597 & .601 & .806 & .687 & .488 & .755  \\ \hline
CE w/o upsample  & .628 & .595 & .599 & .782 & .668 & \textbf{.521} & .695  \\ \hline
\end{tabular}}
\end{table*}

\begin{table}
\footnotesize
\caption{\label{tab:changed_class_weights_res}
AD classification results for the proposed HyperFusion and the late-fusion method. The first row presents the scores obtained our model using different class weight configurations in the weighted cross entropy loss function. The second row presents reported results of the late-fusion method of~\citep{prabhu2022multimodal_AD_entropydecision}. The comparison is based on three metrics detailed in main manuscript.
}
\centering
\begin{tabular}{|l|c|c|c|}
\hline
\textbf{model} & \textbf{BA} & \textbf{PRC} & \textbf{f1 macro} \\ \hline

HyperFusion - different class weights & 0.6539 & 0.6563 & 0.6433 \\ \hline

Reported late-fusion results & 0.6330 & 0.6473 & 0.6240  \\ 
\citep{prabhu2022multimodal_AD_entropydecision}&&&\\
\hline
\end{tabular}
\end{table}

\textcolor{black}{\section{Selection of hyperlayers} \label{sec:hyperlayers-selection}}  
\textcolor{black}{
As detailed in Section~\ref{sec:hyperlayers-selection} in the main paper, there is one hypernetwork embedding for each hyperlayer. The selection of hyperlayer candidates is based on the heuristic suggested in~\cite{lutati2021hyperhyper}. Specifically, we tested how random weight initializations of a specific layer (while keeping all other weights fixed) influences the loss of the overall network. Calculating the entropy of the loss values we evaluate the impact of that layer. After identifying potential candidate layers, fine-tune selection is conducted empirically.}

\begin{figure*}[h]
    \centering
    \includegraphics[scale=0.55]{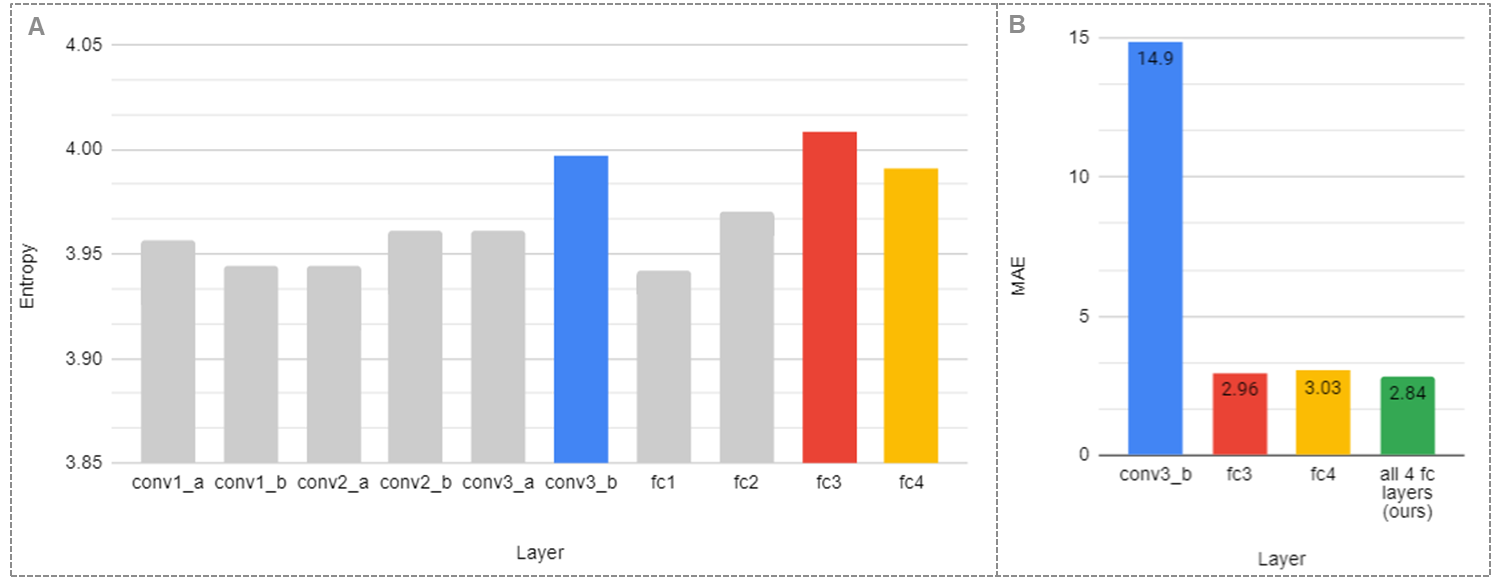}
    \caption{\textcolor{black}{A. A bar plot representing the entropy of the loss values obtained for random initializations of each primary network layer. conv$j$ denotes a convolutional layer within the block (conv1 or 2), and fc$i$ denotes the $i$th fully connected layer. B. A bar plot depicting the ablation study conducted for the selection of hyperlayer combination. Specifically, the MAE for each selection is presented. Both plots refer to brain age prediction.}
    }
    \label{fig:_hyperlayer_brainage_ablation}
\end{figure*}

\begin{figure*}[h]
    \centering
    \includegraphics[scale=0.65]{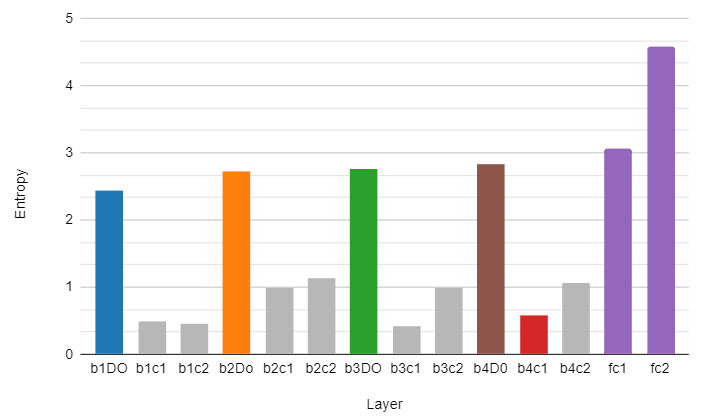}
    \caption{\textcolor{black}{A bar plot representing the entropy of the loss values obtained for random initialization of each primary network layer for AD classification. b$i$ denotes the~$i-$th Res-block, and c$j$ denotes the convolutional layer within the block (conv1 or 2), and fc$i$ denotes the $i$th fully connected layer. Gray bars represent layers that have not been empirically tested.} 
    }
    \label{fig:losses_entropies}
\end{figure*}

\begin{figure*}[h]
    \centering
    \includegraphics[scale=0.65]{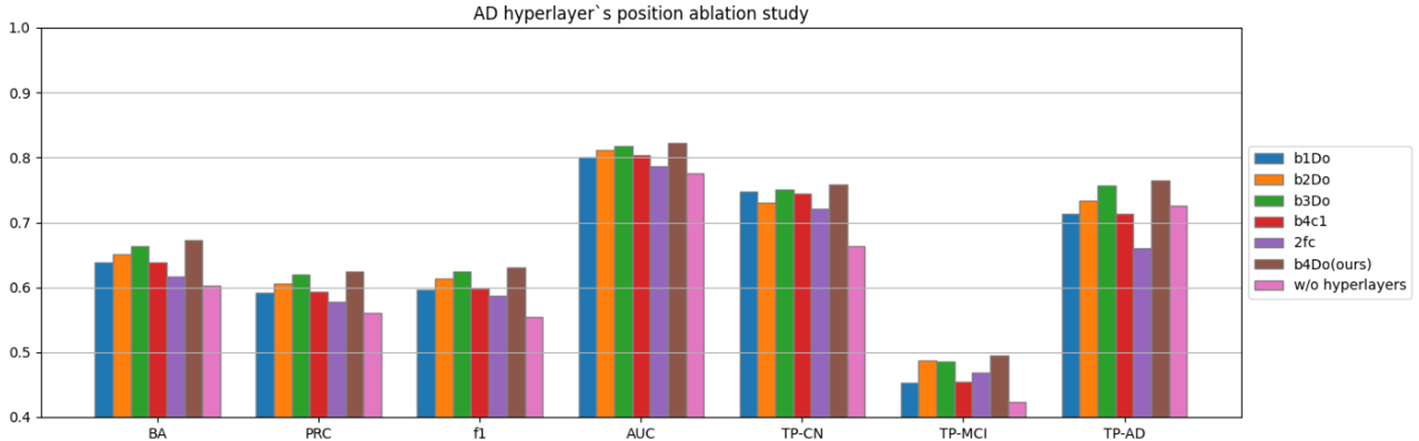}
    \caption{\textcolor{black}{A bar plot depicting the results of the ablation study conducted for the determination of the hyperlayers for AD classification. BA stands for Balance Accuracy; Prc for percision; AUC stands for the Area Under the ROC Curve (AUC) and f1 stands for the F1 score. $TP-C$ represent the True Positive values of class $C$, where , $C\in\{\mbox{CN, MCI, AD}\}$.  The bar colors follow the legend and correspond to these shown in Figure~\ref{fig:losses_entropies}. Legend: DO refers to the downsample convolution inside the block. Other symbols as in~Figure~\ref{fig:losses_entropies}.} \label{fig:diff_hyper_ablation}  }
\end{figure*}

\textcolor{black}{\subsection{Hyperlayers for Brain Age Prediction Network} \label{sec:brainage_hyperlayers-selection}} 
\textcolor{black}{Figure~\ref{fig:_hyperlayer_brainage_ablation}.A presents a bar plot showing the loss entropy for each layer in the primary network. The final combination of hyperlayers is chosen based on an ablation study applied to the selected candidates. Figure~\ref{fig:_hyperlayer_brainage_ablation}.B presents a bar plot showing the prediction results (using the validation set) for each combination tested. Based on this experiment four linear layers in the primary network were set as hyperlayers since this combination yielded the lowest Mean Absolute Error (MAE) for brain age prediction.}\\
\textcolor{black}{\subsection{Hyperlayers for AD classification Network} \label{sec:AD_hyperlayers-selection}} 
\textcolor{black}{Figure~\ref{fig:losses_entropies} depicts the entropy of layer-based loss values for each layer in the primary network for AD classification. Figure~\ref{fig:diff_hyper_ablation} presents a bar plot depicting the results of the ablation study conducted for the final determination of the hyperlayers. Both figures visually demonstrate the correlation between the loss-based entropy of the convolutional layers, their positions within the network, and the resulting performance. Indeed, the linear layers have higher entropy, yet the final prediction results based on the validation set are inferior when these layers were set to be hyperlayers.
In contrast, setting the down-sample convolutional layer in the last Res-block as hyperlayer yielded the best results in all metrics tested (brown bars). This layer was therefore selected.}

\end{document}